\documentclass[10pt,twocolumn,letterpaper]{article}

\usepackage[pagenumbers]{cvpr} 
\usepackage{tabularx}
\usepackage{booktabs}

\definecolor{cvprblue}{rgb}{0.21,0.49,0.74}
\usepackage[pagebackref,breaklinks,colorlinks,allcolors=cvprblue]{hyperref}

\usepackage[marginal]{footmisc}

\def\paperID{5768} 
\def\confName{CVPR}
\def\confYear{2026}

\title{
LoFA: Learning to Predict Personalized Priors for Fast Adaptation of\\
Visual Generative Models
}

\author{
Yiming Hao\textsuperscript{\rm 1}$^{\ref{footnote:equal}}$  \quad Mutian Xu\textsuperscript{\rm 1}$^{\ref{footnote:equal}}$ \quad Chongjie Ye\textsuperscript{\rm 2,3,1} \quad Jie Qin\textsuperscript{\rm 1} \quad Shunlin Lu\textsuperscript{\rm 4} \\
Yipeng Qin\textsuperscript{\rm 5} \quad Xiaoguang Han\textsuperscript{\rm 1,2,3}$^{\ref{footnote:corrs}}$ \vspace{5pt}\\
\normalsize \textsuperscript{\rm 1}{SSE, CUHKSZ} \qquad \textsuperscript{\rm 2}{FNii-Shenzhen} \\
\normalsize \textsuperscript{\rm 3}{Guangdong Provincial Key Laboratory of Future Networks of Intelligence, CUHKSZ} \\
\normalsize \textsuperscript{\rm 4}{SDS, CUHKSZ} \qquad \textsuperscript{\rm 5}{Cardiff University}
}

\begin{document}


\twocolumn[{%
\renewcommand\twocolumn[1][]{#1}%
\maketitle
\begin{center}
    \centering
    \vspace{-2.0em}
    \includegraphics[width=1.0\linewidth]{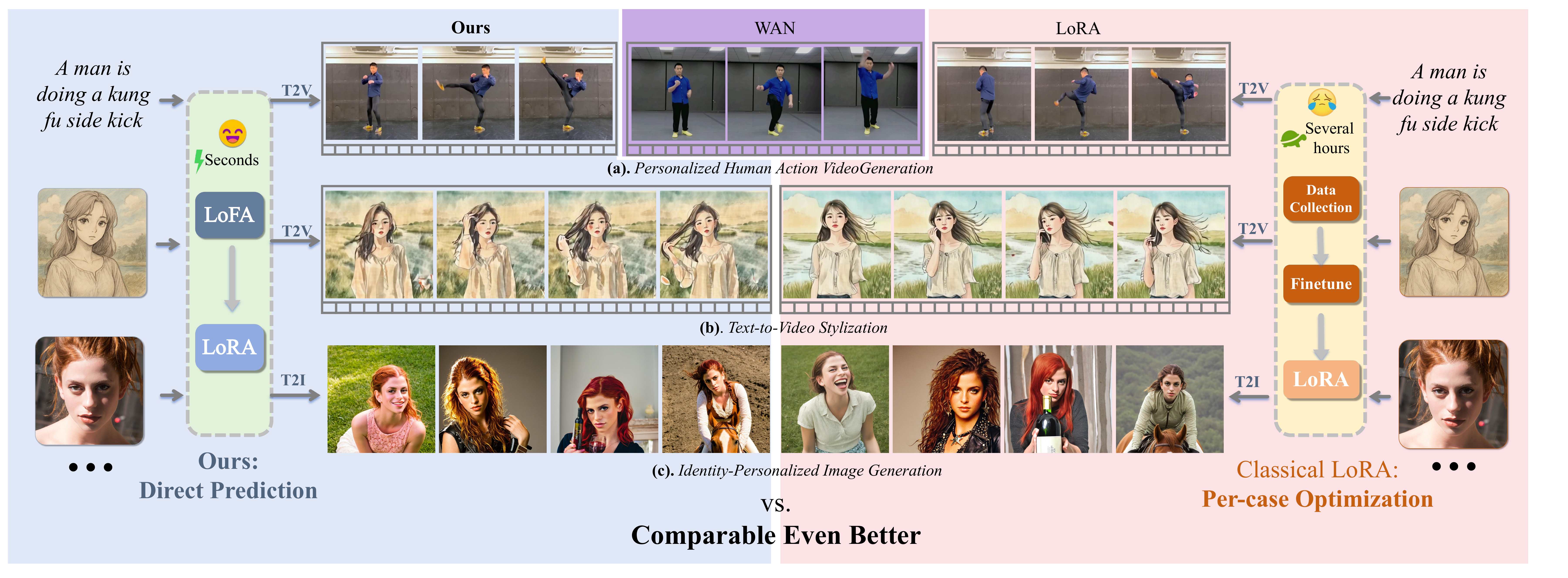}
    \vspace{-1.0em}
    \captionof{figure}{We propose \textbf{LoFA}, a general framework that predicts personalized priors (\ie, LoRA weights) \textbf{within seconds} for fast adaptation of visual generative models. We evaluate its effectiveness across multiple personalization tasks: \textbf{(a)} Personalized Human Action Video Generation, \textbf{(b)} Text-to-Video Stylization, and \textbf{(c)} Identity-Personalized Image Generation. Across all tasks, our LoFA achieves \textbf{comparable or superior generation quality compared to conventional LoRA fine-tuning}—which typically requires hours of data collection and expert optimization. It shows the potential of our LoFA to benefit more practical applications.}
    \label{fig:teaser}
\end{center}
}]

\footnotetext[1]{\label{footnote:equal}Y. Hao and M. Xu contribute equally.}
\footnotetext[2]{\label{footnote:corrs}Corresponding author.}

\begin{abstract}
Personalizing visual generative models to meet specific user needs has gained increasing attention, yet current methods like Low-Rank Adaptation (LoRA) remain impractical due to their demand for task-specific data and lengthy optimization. While a few hypernetwork-based approaches attempt to predict adaptation weights directly, they struggle to map fine-grained user prompts to complex LoRA distributions, limiting their practical applicability.
To bridge this gap, we propose LoFA, a general framework that efficiently predicts personalized priors for fast model adaptation. We first identify a key property of LoRA: structured distribution patterns emerge in the relative changes between LoRA and base model parameters. Building on this, we design a two-stage hypernetwork: first predicting relative distribution patterns that capture key adaptation regions, then using these to guide final LoRA weight prediction. Extensive experiments demonstrate that our method consistently predicts high-quality personalized priors within seconds, across multiple tasks and user prompts, even outperforming conventional LoRA that requires hours of processing. Project page: \href{https://jaeger416.github.io/lofa/}{jaeger416.github.io/lofa}.
\end{abstract}    
\section{Introduction}
\label{sec:intro}

In recent years, the growing demand for creative media and visual content \cite{visaulart} has driven the development of powerful visual generative foundation models \cite{stablevideodiffusion,saharia2022photorealistic,stablediffusion,wan2025wan}. 
Trained on large-scale image or video datasets, these models exhibit rich capabilities and general prior knowledge across diverse domains.
However, when faced with users' personalized needs—particularly those involving fine-grained prompts—they often struggle to generate outputs that closely align with user intent (\eg, as shown in \cref{fig:teaser}-“WAN Original Output”, the text-to-video foundation model \cite{wan2025wan} struggles to capture the personalized motion prompt like `\textit{a man is doing a kung fu side kick}').

To address this issue, early studies typically employ parameter-efficient fine-tuning techniques \cite{dong2022dreamartist,textualinversion,ruiz2023dreambooth,customdiffusion} to adapt models with personalized prior knowledge. However, these methods require optimizing a separate adapter (\eg, Low-Rank Adaptation (LoRA) \cite{hu2022lora}) for each individual personalization, demanding both task-specific data and substantial optimization time (see `classical LoRA' in \cref{fig:teaser}). This makes them impractical for real-world applications where users expect fast adaptation to novel queries.

To enable fast adaptation, a limited number of recent works \cite{hyperdreambooth,wu2024difflora} have attempted to directly predict adaptation weights at test time. For example, Ruiz \etal \cite{hyperdreambooth} introduced a hypernetwork-based approach for this purpose, though it still requires additional post-optimization. Wu \etal \cite{wu2024difflora} build on this direction by removing post-optimization entirely. However, this solution path faces a \textit{fundamental challenge}: the model must learn a complex mapping from \textit{low-dimensional, fine-grained} user prompts to \textit{high-dimensional, complex} LoRA weight distributions. Due to this difficulty, existing approaches have only been validated for subject identity personalization in image generation—a relatively constrained scenario. This limitation likely stems from their use of LoRA weight compression into lower-dimensional spaces as hypernetwork outputs, which inevitably causes information loss and restricts model capacity. Consequently, a viable/versatile solution for fast model adaptation that can effectively handle fine-grained user prompts or high-dimensional, complex LoRA outputs—essential capabilities for practical user-centric applications—remains an open and significantly underexplored research challenge in visual generation.


To bridge this gap, we propose LoFA, a general learning framework that is able to directly predict personalized priors from diverse or fine-grained user prompts for fast adaptation of visual generative models (see `LoFA' in \cref{fig:teaser}). 
Our core idea lies in embedding novel \textit{guidance} into the hypernetwork design, enabling it to predict complete, \textit{un}compressed LoRA weights directly from user prompts without resorting to lossy compression.
To achieve this, we first identify a key property of LoRA, denoted as \textit{response maps}, which are distinct structured patterns emerge in the \textit{relative} changes between personalized LoRA weights and their original model parameters, capturing the primary influence of user prompts (visualized in \cref{fig:response_map} and analyzed in \cref{sec:finding_lora_response}).
Building on this finding, we design a new architecture that avoids brute-force prompt-to-LoRA mapping. Instead, it takes original base model weights as input and integrates user prompts through cross-attention to learn relative adaptations.
This learning process is further split into two stages, where the network first predicts response maps—significantly lower-dimensional and simpler than LoRA weights. Next, the learned response knowledge then guides the final LoRA prediction to recognize and prioritize key adaptation regions, helping to ease and stabilize learning. This design enables the network to learn \textit{relative adaptations} between base models and target LoRAs through \textit{structured response guidance}, while predicting complete and \textit{un}compressed LoRA weights with full expressive power.

We verify the effectiveness of our framework under various types of user-specific inputs, including text, pose, style reference, and human face, for both image and video generation tasks. Our method not only substantially outperforms baseline approaches but also achieves performance comparable to—and in many cases superior to—individually optimized LoRAs, demonstrating the viability of fast model adaptation for practical applications.

Our contributions are as follows: 
\begin{itemize}
    \item We propose LoFA, a novel and versatile framework for fast adaptation of visual generative models that directly predicts personalized priors from diverse user prompts.
    \item We identify and formally characterize the distinct structures in LoRA response distributions, and propose the first method that leverages this property to guide personalized LoRA prediction.
    \item We conduct extensive experiments across multiple visual generation tasks using different user prompts, demonstrating strong performance on practical applications.
    \item We will release code and pre-trained models to support reproducibility and future research.
\end{itemize}
\section{Related Work}
\label{sec:related_work}

\paragraph{Parameter-Efficient Fine-Tuning (PEFT)} enables the adaptation of large foundation models to specialized domains \cite{dontstoppretraining2020,wei2021finetuned}. Early approaches \cite{rebuffi2017learning,houlsby2019parameter,lin2020exploring} introduced adapter layers inserted between existing network layers. Recently, the field advanced significantly with low-rank adaptation methods that merge with original model weights, addressing efficiency bottlenecks \cite{karimi2021compacter,hu2022lora}, where LoRA \cite{hu2022lora} has emerged as the most representative and widely used approach. AdaLoRA \cite{zhang2023adaptive} and VeRA \cite{kopiczko2024vera} present more lightweight LoRA variants through further matrix pruning or decomposition. In our work, we adopt LoRA as the output weight adapter since it preserves maximum model expressiveness—particularly crucial for adapting complex, high-dimensional models such as video generative models.

\paragraph{Personalization of visual generative models} aims to produce visual contents that reflect specific user-provided references, such as particular objects, styles, or motion patterns. Early efforts primarily relied on Generative Adversarial Networks (GANs), with methods like StyleGAN \cite{karras2020analyzing} enabling control over visual attributes by manipulating latent spaces. Recently, the advent of diffusion models \cite{ho2020denoising} catalyzed a new path of designing modules to directly incorporate user prompts. IP-Adapter \cite{ye2023ip} and ControlNet \cite{zhang2023adding} integrate reference conditions via additional feature injection for image stylization, while \cite{singer2022make,das,shi2024motion,ren2025gen3c,zhang2025recapture,mark2025trajectorycrafter} leverage temporal adaptation or 3D-aware rendering given subject or camera motion for controllable video synthesis.
However, to acquire personalized priors, these methods typically require fine-tuning parts of the base foundation model or training additional adapters using PEFT techniques, as explored in \cite{dong2022dreamartist,textualinversion,ruiz2023dreambooth,customdiffusion}, demanding both user-specific data and considerable training time. This renders them imperfect for real applications where users expect fast adaptation to new queries. To overcome this, we propose a general learning framework capable of directly predicting personalized priors from novel user prompts without any retraining.

\paragraph{Hypernetworks for predicting adaptation weights.}
HyperNetworks \cite{ha2016hypernetworks} and its variants \cite{peebles2022learning,wang2024neural,wangscaling} explore using one network to generate weight parameters of another neural network. 
A handful of recent works have adopted this paradigm to generate LoRA parameters for fast model adaptation. Text-to-LoRA \cite{charakorn2025texttolora} produces task-specific LoRAs for NLP tasks, while \cite{hyperdreambooth,wu2024difflora,jin2024conditional,lorarar,incontext} apply similar frameworks to visual foundation models. However, these methods all compress LoRA weights via autoencoders or decomposition to simplify learning—inevitably causing information loss and limiting model capacity. While adequate for subject identity or style adaptation (relatively limited toy scenarios), they underperform on fine-grained/user-specific inputs and high-dimensional/complex LoRA outputs needed for practical applications. Fast model adaptation from user-specific prompts thus remains underexplored. To address this, we propose a new LoRA hypernetwork with inductive biases to ease learning while preserving full LoRA expressiveness—without compression.
\section{Method}

\subsection{Preliminary and Overview}

\paragraph{Preliminaries of LoRA.}
Low Rank Adaptation (LoRA)~\cite{hu2022lora} has emerged as a representative Parameter-Efficient Fine-Tuning (PEFT) technique by introducing trainable low rank weight residuals alongside frozen layers. Specifically, for a layer $l$ with weight matrix  $W \in \mathbb{R}^{m \times n}$, LoRA only updates its residual weight $\Delta W$, which is further decomposed into two low rank matrices $B \in \mathbb{R}^{m \times r}$ and 
$A \in \mathbb{R}^{r \times n}$, where $r \ll \min(m,n)$. The final adapted weights can be represented as:
\begin{equation}
W' = W + \Delta W = W + B A.
\label{eq:lora}
\end{equation}

\paragraph{Framework overview.}
During training, given task-specific pairs of user prompt condition $C$ and target LoRA weights $\Delta W$, along with the base visual foundation model weight $W$, our framework learns to predict $\Delta W$ from $C$ and $W$, using a two-stage transformer-based \cite{attention} hypernetwork $f$, which can be formulated as: 
\begin{equation}
    f(C, W) \rightarrow \Delta W.
\end{equation}
Specifically, we first identify the structured pattern in LoRA response, reflected by the ratio map between LoRA weight and base model weight (\cref{sec:finding_lora_response}). Based on this finding, as shown in \cref{fig:framework} and \cref{sec:main_framework}, our network first predicts a relatively simple, low-dimensional response map. Next, the knowledge acquired in this initial stage then guides the final prediction of the complete LoRA weight.

\begin{figure}[t]
  \centering
  \includegraphics[width=0.8\linewidth]{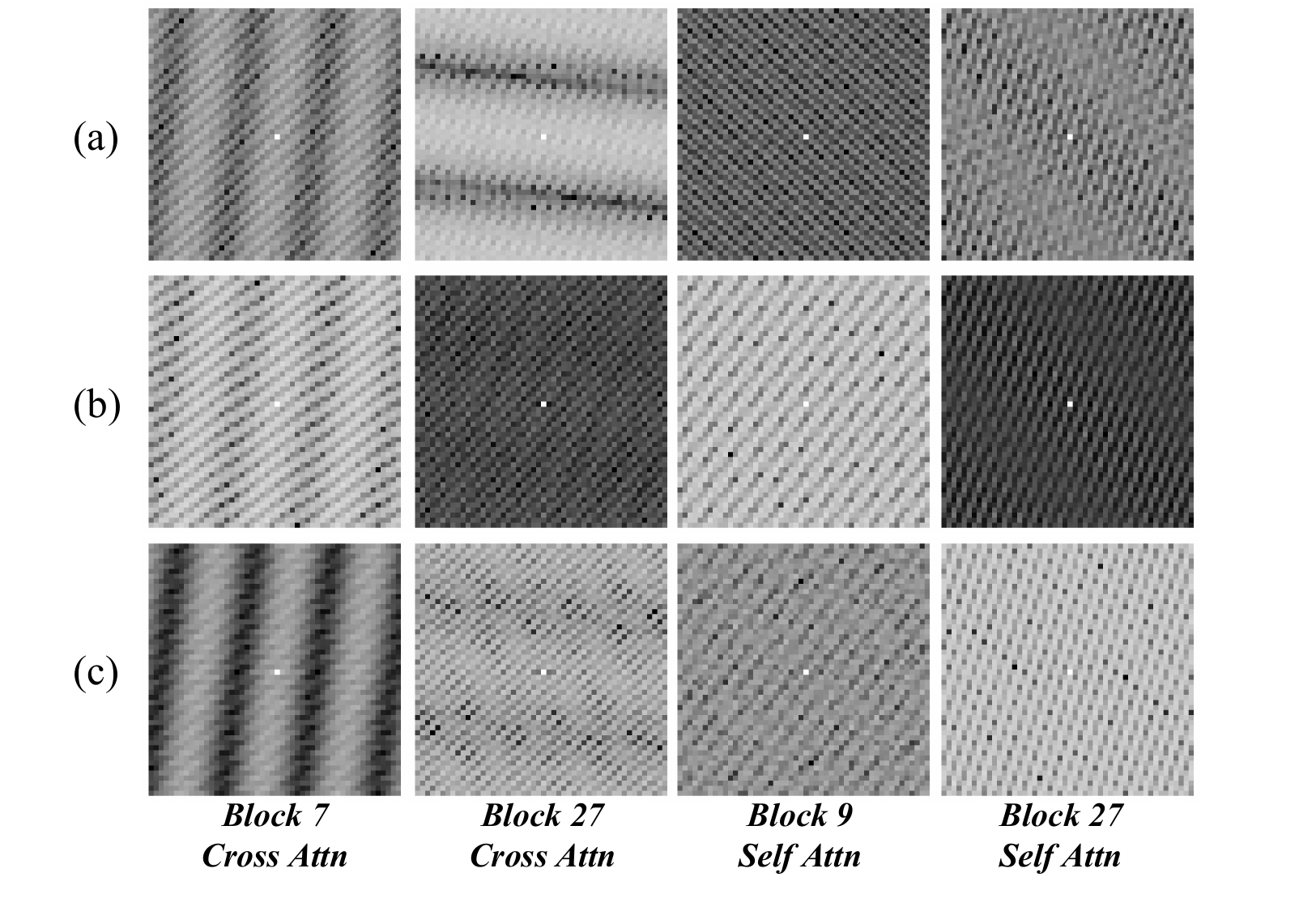}
  \caption{\textbf{Visualization of LoRA response maps}. Each row corresponds to a distinct task-specific LoRA, while columns represent different network layers or blocks.}
  \label{fig:response_map}
\end{figure}

\subsection{Key Finding on LoRA Distribution Pattern}
\label{sec:finding_lora_response}

\paragraph{\textit{Revisiting} the distribution explorations of LoRA.}
The high dimensionality of LoRA makes direct full-weight prediction difficult, particularly in video generative models with an additional temporal dimension.
To address this, existing methods try to leverage the \textit{sparse distribution} of LoRA, straightforwardly reducing the LoRA parameters into a lower-dimensional space using auto-encoders \cite{wu2024difflora,wangscaling,jin2024conditional}, matrix decomposition \cite{kopiczko2024vera,hyperdreambooth}, or pruning \cite{zhang2023adaptive}. However, the brute-force reduction of parameters inevitably introduces information loss and constrains model capacity.

\begin{figure*}[t]
    \centering
    \includegraphics[width=1.0\linewidth]{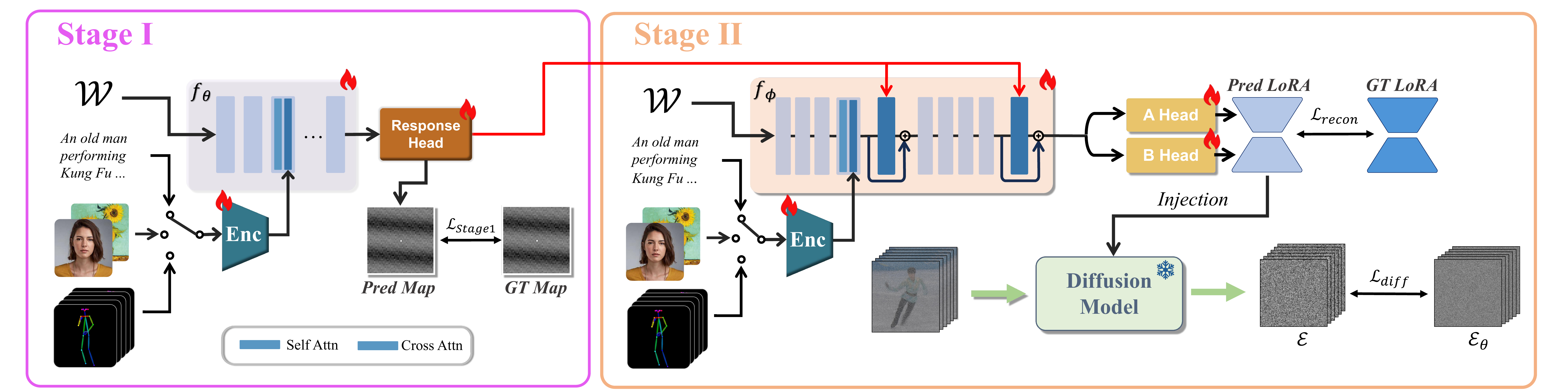}
    \caption{\textbf{An overview of our LoFA.} Conditioned on different user prompts, our network takes the base model weight $W$ as the input, and predicts LoRA response map ${\hat R}$ (\cref{fig:response_map}) at Stage-I. Next, Stage-II inherits Stage-I's architecture, and uses the learned information of the response map to guide the final prediction of the full LoRA weights.}
    \label{fig:framework}
\end{figure*}

\paragraph{\textit{Finding} the structured pattern in the relative changes between LoRA and base model weights.}
To this end, reflecting on the residual nature of LoRA (see \cref{eq:lora}: LoRA is a $\Delta$ on base model weight), we start by taking a closer look at the ratio map (\ie, indicating relative changes) between LoRA weights $\Delta W$ and the base model weights $W$.
Statistically, for the $i$-th parameter in a layer, we define its normalized response magnitude as $m_i = |\Delta w_i / w_i|$, where $\Delta w_i \in \Delta W$ and $w_i \in W$. We find that most responses (approximately $50\%\text{--}80\%$, depending on the layer) are below $2\%$.
Based on this, we apply a $2\%$ threshold (ablated in the supplementary material) to get a binary-masked response distribution map: $R = \{ r_i|r_i = 1 \text{ if } m_i>2\% \text{ else } 0 \}$, where $r_i=1$ and $r_i=0$ respectively denote useful and negligible parameters. \cref{fig:response_map} visualizes $R$, revealing distinct distributions across different task specifications.

\paragraph{\textit{Utilizing} the response pattern as guidance.}
The observed parameter response distribution reveals which parts of the base model are most significantly adapted by LoRA, capturing the primary effects of task conditions. 
If we can find a way to \textit{softly} inject this property into the hypernetwork—guiding it to recognize and prioritize key adaptation regions—we will simplify and stabilize the learning process. Moreover, the full number of the original LoRA parameters could be preserved, maintaining their expressive power, while enabling the network to intrinsically/implicitly learn the distinct structure of the adaptation.

\subsection{Main Framework: Two-stage Prediction}
\label{sec:main_framework}

\paragraph{Stage-I: LoRA response prediction.}
Building upon the above observations, the first stage of our framework is a transformer-based \cite{attention} hypernetwork $f_{\theta}$ that takes the base model’s original parameters $W$ as input and predicts the parameter response map ${R}$.
Within the Transformer, self-attention modules capture the implicit correlations among parameters, while cross-attention integrates the user prompt $C$ as a conditioning signal.
Moreover, as we find that ${R}$ varies across different depth and block function types (\eg, query projections, key projections, \etc) from our visualizations (\cref{fig:response_map}), we regard each ${R}$ from different depth or blocks as one sample.
We further introduce two learnable embeddings: \emph{blockwise position embedding} $E_{\text{pos}}$ encodes the layer depth index of each transformer block and \emph{block-type embedding} $E_{\text{type}}$ represents the functional role of the type. The learning process is formulated as:

\begin{equation}
f_{\theta}\!\left(W+E_{\text{pos}}+E_{\text{type}}, C\right)
\rightarrow \hat{R},
\end{equation}
where $\hat{R}$ is the predicted response map, activated by the Sigmoid function.
The first stage is trained with a standard cross-entropy loss, defined as:
\begin{equation}
\mathcal{L}_{\text{stage1}} 
= -R \, \log \hat{R}.
\end{equation}

\paragraph{Stage-II: LoRA weight prediction.}
The Stage-II model $f_{\phi}$ retains the transformer backbone from the first stage and replaces the MLP head with two separate parameter prediction heads to predict the LoRA parameter matrices $B$ and $A$, for different layers and blocks, summed to get the final LoRA weights $\Delta W = BA$. The transformer is initialized from the Stage-I model to preserve the learned response priors and parameter dependencies.
Moreover, we introduce additional cross-attention layers in specific blocks, which attend to the final-layer feature representation $F_{\text{stage1}}$ of the Stage-I model.
Formally, the Stage-II learning can be expressed as:
\begin{equation}
f_{\phi}\!\left(
W +E_{\text{pos}} + E_{\text{type}}, 
C, 
F_\text{stage1}
\right) \rightarrow \big[\hat{B}, \hat{A}\big] ,
\end{equation}
Following~\cite{hyperdreambooth}, the second stage is jointly optimized with two complementary objectives: a reconstruction loss $\mathcal{L}_{\text{recon}}$ and a diffusion loss $\mathcal{L}_{\text{diff}}$. 
$\mathcal{L}_{\text{recon}}$ is an $L_1$ objective between the predicted and ground-truth LoRA parameters:

\begin{equation}
\mathcal{L}_{\text{recon}} =
\lVert A - \hat{A} \rVert_1
+ \lVert B - \hat{B} \rVert_1.
\end{equation}

$\mathcal{L}_{\text{diff}}$ is computed by injecting all predicted LoRAs into the base denoising model, and optimized under the Flow Matching paradigm \cite{lipman2022flow}. It provides task-level supervision by guiding the hyper-network toward generating LoRA adapters that produce desirable diffusion outputs consistent with the target generative behavior, formulated as:
\begin{equation}
\mathcal{L}_{\text{diff}} = 
\mathbb{E}_{x_0, t, \epsilon}
\big[
\| \epsilon - \epsilon_\theta(x_t, t; \hat{A}, \hat{B}) \|_2^2
\big],
\end{equation}
where $x_0$ is a clean sample, $t$ and $\epsilon$ respectively denote the added Gaussian noise and the time step during the forward diffusion process, and $\epsilon_\theta$ indicates the noise prediction network.
The overall training objective is a weighted sum of the two terms:
\begin{equation}
\mathcal{L}_{\text{stage2}} =
\lambda_{\text{recon}} \mathcal{L}_{\text{recon}} 
+ 
\lambda_{\text{diff}} \mathcal{L}_{\text{diff}}.
\end{equation}
This dual-objective formulation ensures that the predicted LoRA parameters are both structurally plausible and functionally aligned with the diffusion objective.

\paragraph{Downstream inference.} 
In downstream applications, when a novel task-specific prompt $C$ is provided, our trained hypernetwork $f$ directly predicts the corresponding LoRA weight $\Delta W$ in a short time, following \cref{eq:lora}. This enables fast model adaptation of the base model into a newly personalized version, effectively incorporating personalized priors that closely reflect the user prompt.
\section{Experiments}
We conduct comprehensive experiments to evaluate the effectiveness of our LoFA framework across both video and image generation tasks. To demonstrate its versatility in handling diverse prompt conditions, we test multiple input modalities across three key applications.
For video generation, we use WAN2.1‑1.3B \cite{wan2025wan} as the base model and evaluate on:
(1) \textbf{Personalized human action video generation} from text or motion poses (\cref{sec:exp_action_video}), addressing the challenging personalization of dynamic motion—a core attribute of video data;
(2) \textbf{Text-to-video stylization} using style images as reference (\cref{sec:exp_video_style}), a classic task in video editing.
For image generation, we adopt Stable Diffusion XL \cite{podell2024sdxl} and evaluate on:
(3) \textbf{Identity-personalized image generation} (\cref{sec:exp_id_image})—the only scenario previously supported by related works \cite{hyperdreambooth,wu2024difflora}.

A user study, more qualitative results and implementation details can be found in the supplementary material.


\subsection{Personalized Human Action Video Generation}
\label{sec:exp_action_video}

For this task, we employ two distinct forms of personalized conditioning: textual prompts and 2D pose sequences.

\paragraph{Preparation and setups.}
For text-conditioned generation, we select 39.2k single-human-action videos from MotionX \cite{lin2023motion} and MotionX++ \cite{zhang2025motion}. Each video is re-annotated with a descriptive text prompt using Qwen2.5-VL-7B-Instruct \cite{bai2025qwen2}. Based on motion semantics from the dataset, we group videos with similar action labels into sub-task-specific training sets, each corresponding to a distinct LoRA (\eg, one LoRA for yoga videos, another for dancing videos, \etc). Within each group, video–text pairs capture varied motion details (\eg, within yoga: `pull-ups', `diagonal arm push', `straddle side stretch'). In total, we train 2,630 LoRAs, forming a text–LoRA dataset built with moderate effort. For evaluation, we sample 60 unseen action types (\eg, kungfu, fistfight—beyond yoga or dancing), yielding 877 text–LoRA pairs as a held-out validation set.

For pose-conditioned generation, we reuse the same video set and extract 2D skeletal keypoint sequences using DWPose \cite{yang2023effective}. Following the ControlNet pipeline \cite{zhang2023adding}, we render these poses into RGB sequences as structural conditioning signals. As in UniAnimate \cite{wang2025unianimate}, we use a multi-layer 3D convolutional network as the pose encoder to extract spatio-temporal motion representations for guiding LoRA prediction. 

\begin{table}[t]
  \centering
  \resizebox{1.0\linewidth}{!}{
  \begin{tabular}{lccc}
    \toprule
    Method & FVD $\downarrow$ & CLIP-T $\uparrow$ & Dynamic~Degree $\uparrow$ \\
    \midrule
    \small{\textit{Per-case Optimization}}\\
    LoRA \cite{hu2022lora} & 609.5 & 0.3662 & 0.2269 \\
    \midrule
    \small{\textit{Direct Prediction}}\\
    Text-to-LoRA \cite{charakorn2025texttolora} & 907.5 & 0.3541 & 0.0745 \\
    LoFA-Text Cond. (\textbf{Ours}) & \textbf{589.8} & \textbf{0.3719} & 0.2283 \\
    LoFA-Pose Cond. (\textbf{Ours}) & 610.7 & 0.3687 & \textbf{0.2297} \\
    \bottomrule
  \end{tabular}
  }
  \caption{
  Quantitative comparison on text/pose-conditioned \textbf{Personalized Human Action Video Generation}. 
  }
  \label{tab:main_metrics}
\end{table}

\begin{figure*}[t]
    \centering
    \includegraphics[width=1.0\linewidth]{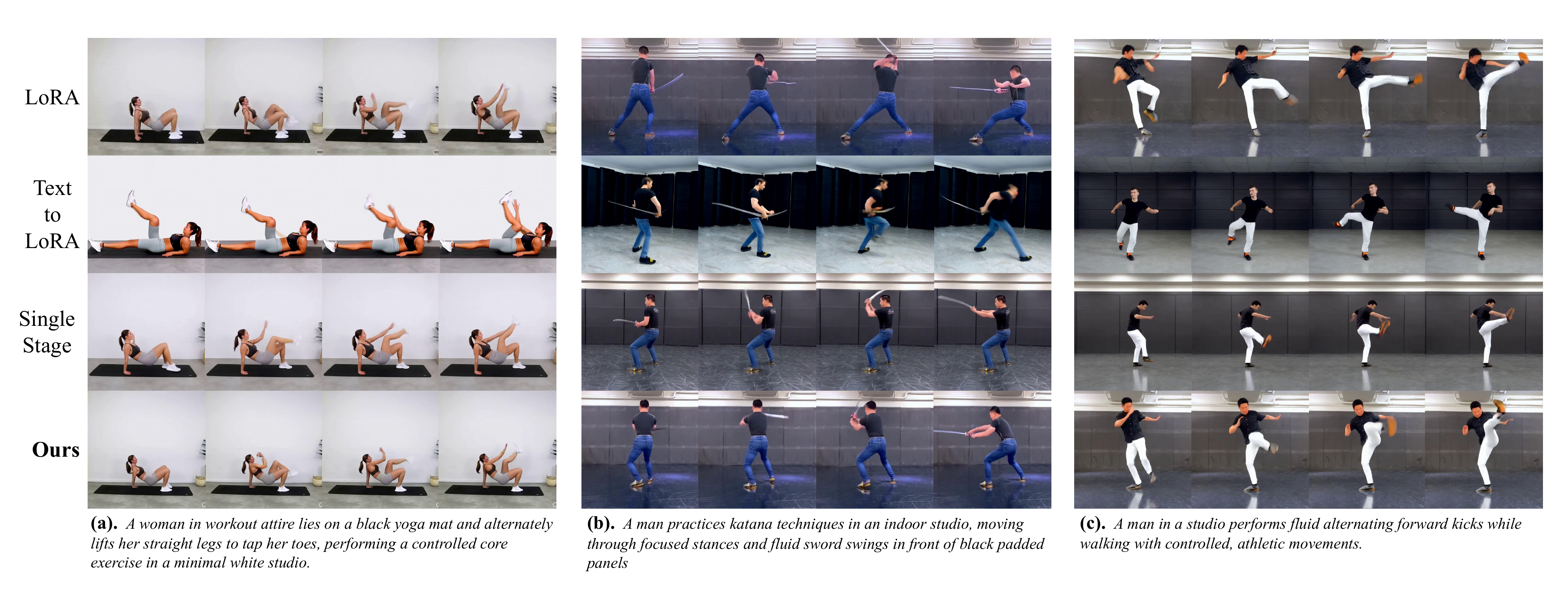}
    \caption{Qualitative results on \textbf{text-conditioned Personalized Human Action Video Generation}.}
    \label{fig:text_cond_result}
\end{figure*}

\begin{figure*}[t]
    \centering
    \includegraphics[width=1.0\linewidth]{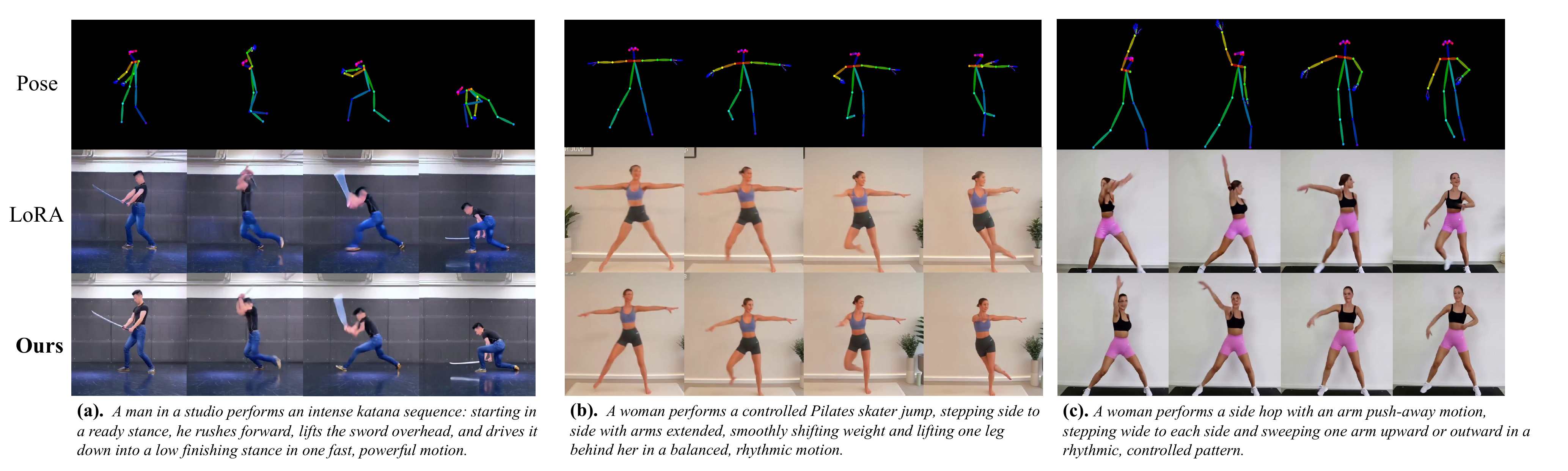}
    \caption{Qualitative results on \textbf{pose-conditioned Personalized Human Action Video Generation}.}
    \label{fig:pose_cond_result}
\end{figure*}

\paragraph{Baseline methods.}
Among existing published works, HyperDreamBooth \cite{hyperdreambooth} is the only conceptually related approach, but lacks officially released code. Moreover, adapting it to video generation is impractical, as its single-network prediction of all LoRA parameters (rather than layer/block-wise) would lead to parameter explosion.
We therefore use Text-to-LoRA \cite{charakorn2025texttolora}—originally designed for large language models—as our primary baseline. It offers open-source implementation $^{\ref{footnote:t2l_repo}}$ and native text-conditioned support. For fair comparison, we retrain it on our dataset using its original architecture and training protocol.

\footnotetext[1]{\label{footnote:t2l_repo}\url{https://github.com/SakanaAI/text-to-lora}}

\paragraph{Results.}
Since the core objective of personalized LoRA adaptation is to capture user-specific requirements, we employ several metrics that specifically assess whether generated videos align with personalized conditions.
We first evaluate generated videos using Fréchet Video Distance (FVD) \cite{unterthiner2018towards} to measure distributional similarity to ground-truth videos—which inherently satisfy personalized prompts by definition. We also report CLIP-T scores \cite{hessel2021clipscore} to assess text-video alignment. Given our focus on human action generation, we further employ Dynamic Degree from VBench \cite{huang2023vbench} to quantify dynamic quality.

As shown in \cref{tab:main_metrics}, our method consistently outperforms baseline approaches across all evaluation metrics. Notably, it achieves competitive or superior performance compared to per-case optimized LoRA \cite{hu2022lora}. We attribute this to the observation that original LoRAs are fine-tuned on individual subtasks, which may cause mild overfitting to their respective training data. In contrast, LoFA is trained across diverse subtasks, learning more generalized motion priors that lead to higher-quality video generation.
We also observe that our pose-conditioned model yields slightly lower text alignment and overall video quality than the text-conditioned variant. This likely stems from the text model's use of the T5-XXL encoder \cite{chung2024scaling}, which provides stronger semantic features and richer video detail representations. Nevertheless, the pose-based model delivers competitive Dynamic Degree, as the explicit pose sequences offer effective motion priors.
The qualitative results are presented in \cref{fig:text_cond_result} and \cref{fig:pose_cond_result}, where our method shows decent alignment with personalized conditions.

\subsection{Text-to-Video Stylization}
\label{sec:exp_video_style}

\begin{figure*}[t]
    \centering
    \includegraphics[width=1.0\linewidth]{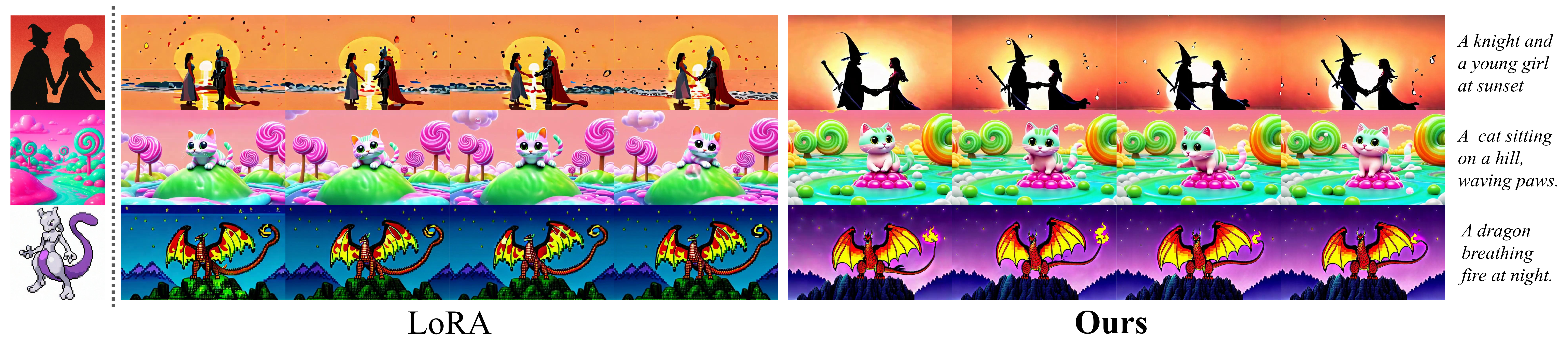}
    \vspace{-0.6cm}
    \caption{Qualitative results on \textbf{Text-to-Video Stylization}.}
    \label{fig:t2v_stylization}
\end{figure*}

\paragraph{Preparation and setups.}
We first collect 1,500 style reference images from the internet. Each image defines a style subgroup, for which we randomly sample 60 text prompts from OpenVid \cite{nan2024openvid}. Using StyleMaster \cite{ye2025stylemaster}, we synthesize stylized videos from each image-text pair, then train a dedicated LoRA for each style subgroup. This process yields 1,500 image-LoRA training pairs, with 100 pairs held out for testing.
The LoRA training setup remains consistent with \cref{sec:exp_action_video}.
The CLIP-ViT-L \cite{radford2021learning} feature of each style image is projected through a learnable MLP to transform it into the conditioning space for LoRA prediction.

\paragraph{Baseline methods.}
Similar to the setup in \cref{sec:exp_action_video}, adapting existing hypernetwork-based methods to video generation remains impractical due to parameter explosion. Furthermore, Text-to-LoRA \cite{charakorn2025texttolora} (used in \cref{sec:exp_action_video}) cannot handle image-conditioned LoRA prediction. We therefore compare only against per-case optimized LoRA \cite{hu2022lora} here.

\begin{table}[t]
  \centering
  \resizebox{1.0\linewidth}{!}{
  \begin{tabular}{lcccc}
    \toprule
    Method & CSD-Score $\uparrow$ & CLIP-T $\uparrow$ & D.D. $\uparrow$ & M.S. $\uparrow$ \\
    \midrule
    \small{\textit{Per-case Optimization}}\\
    LoRA \cite{hu2022lora} & 0.419 & 0.2849 & 2.107 & 0.9816 \\
    \midrule
    \small{\textit{Direct Prediction}}\\
    LoFA (\textbf{Ours})  & \textbf{0.427} & \textbf{0.2943} & \textbf{2.394} & \textbf{0.9940} \\
    \bottomrule
  \end{tabular}
  }
  \caption{
  Quantitative comparison on \textbf{Text-to-Video Stylization}. `D.D.' and `M.S' respectively denote Dynamic Degree and Motion Smoothness from \cite{huang2023vbench}.
  }
  \vspace{-0.2cm}
  \label{tab:style_metrics}
\end{table}

\begin{figure*}[t]
  \centering
  \includegraphics[width=0.9\linewidth]{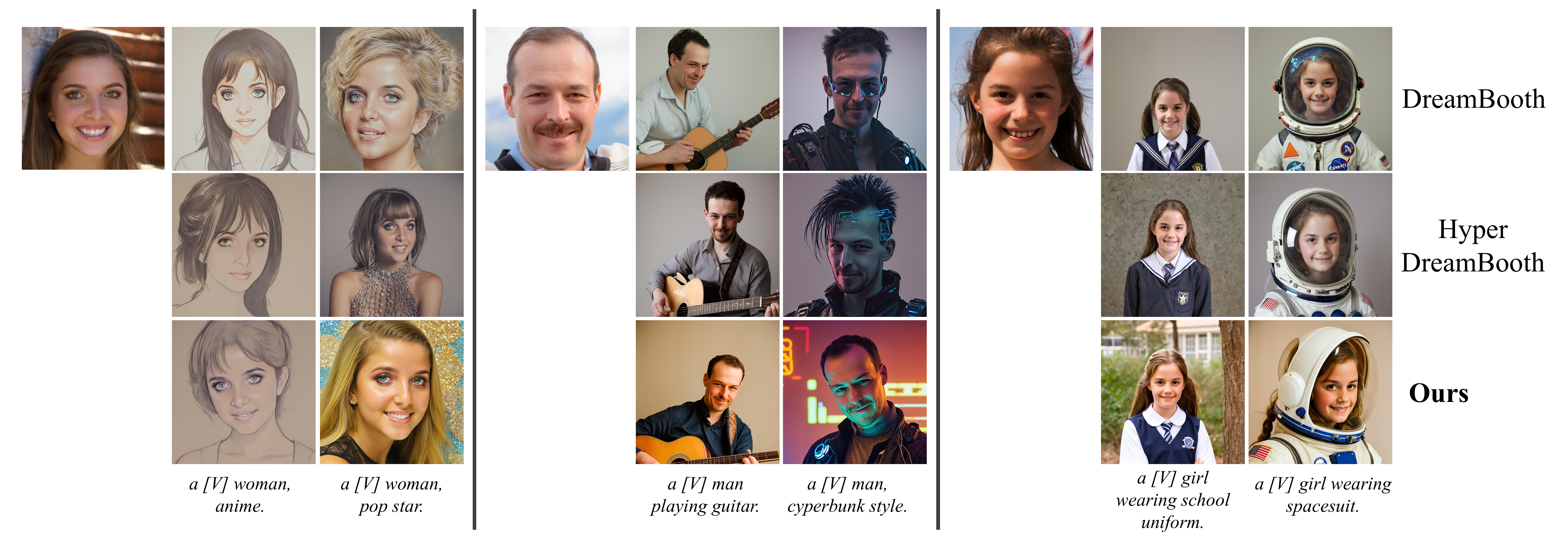}
  \vspace{-0.2cm}
  \caption{Qualitative results on \textbf{Identity-Personalized Image Generation}.}
  \label{fig:face_compare}
\end{figure*}

\paragraph{Results.}
Following \cite{ye2025stylemaster}, we adopt a two-fold evaluation strategy: frame-level stylization quality using image metrics (CSD score \cite{somepalli2024measuring}), and motion quality assessment via VBench \cite{huang2023vbench} metrics (dynamic degree and motion smoothness). Text-video alignment is measured by CLIP-Text \cite{radford2021learning} similarity. \cref{tab:style_metrics} shows that our predicted LoRAs outperform per-case optimized LoRAs across all metrics, with particularly notable gains in dynamic degree. \cref{fig:t2v_stylization} presents qualitative results, showing consistent conclusions.

\subsection{Identity-Personalized Image Generation}
\label{sec:exp_id_image}

\paragraph{Preparation and setups.}
We randomly sample 3,000 face images from FFHQ \cite{karras2019style} as references. 
Following the procedure described in \cite{wu2024difflora}, we use PhotoMaker \cite{li2024photomaker} to generate 86 images for each identity, and then correspondingly train 3,100 LoRAs for each identity using DreamBooth \cite{ruiz2023dreambooth} (a per-case optimized LoRA adaptation method for identity-personalized image generation).
In our LoFA, the CLIP-ViT-L \cite{radford2021learning} feature of each face reference image is projected through an MLP and injected into the transformer as the conditioning signal. 

\paragraph{Baseline methods.}
There exist two relevant hypernetwork-based methods that share conceptual similarities with our approach — HyperDreamBooth \cite{hyperdreambooth} and DiffLoRA \cite{wu2024difflora} — and both are specially designed for identity-personalized image generation. As neither provides official code, we carefully reimplemented both approaches and trained them on our dataset using the configurations specified in their respective papers.

\begin{table}[t]
  \centering
  \resizebox{1.0\linewidth}{!}{
  \begin{tabular}{lcccc|c}
    \toprule
    Method & Face~Sim~$\uparrow$ & DINO~$\uparrow$ & CLIP-I~$\uparrow$ & Face~Div~$\uparrow$ & Time~$\downarrow$ \\
    \midrule
    \small{\textit{Per-case Optimized LoRA}}\\
    DreamBooth \cite{ruiz2023dreambooth}        & 0.488 & 0.460 & 0.544 & 47.3 & 1h \\
    \midrule
    \small{\textit{Direct LoRA Prediction}}\\
    DiffLoRA \cite{wu2024difflora}  & 0.461 & 0.427 & 0.517 & 46.8 & 20s \\
    HyperDreamBooth \cite{hyperdreambooth}   & 0.527 & 0.462 & 0.565 & 46.1 & 274s \\
    LoFA (\textbf{Ours})      & \textbf{0.548} & \textbf{0.497} & \textbf{0.600} & \textbf{50.3} & \textbf{3.7s} \\
    \bottomrule
  \end{tabular}
  }
  \caption{
  Quantitative comparison on \textbf{Identity-Personalized Image Generation}.  
  `Time' denotes forwarding time.
  }
  \label{tab:face_metrics}
\end{table}

\paragraph{Results.}
Following \cite{li2024photomaker}, we evaluate identity preservation using CLIP-I \cite{radford2021learning}, DINO \cite{caron2021emerging}, and Face Similarity metrics, while assessing face diversity through Face Diversity scores. As shown in \cref{tab:face_metrics}, our method achieves superior performance across all evaluation metrics while demonstrating better time efficiency, especially than per-case optimized DreamBooth \cite{ruiz2023dreambooth}, or HyperDreamBooth \cite{hyperdreambooth} that requires post-optimization—even without accounting for data collection time in the comparison. \cref{fig:face_compare} presents qualitative comparisons, showing that our generated images achieve closer visual alignment with the reference images.

\subsection{Ablation Studies and Analysis}
To extensively assess the effectiveness and necessity of key components in our LoFA, we conduct a series of ablation studies and analyses.

\paragraph{LoRA response prediction.}
Since LoRA response prediction at the first stage is the key design of our method, we ablate it across \textit{all} tasks. 
As shown in \cref{tab:ablation} “w/o res”, when the first stage of LoRA response prediction is removed and directly predict LoRA parameters in a single stage, the generation quality degrades significantly across all tasks, especially in more challenging video generation settings. 
This indicates that the LoRA response distribution provides crucial guidance for the final LoRA prediction. 
\begin{table}[t]
  \centering
  \resizebox{1.0\linewidth}{!}{
  \begin{tabular}{c||cc||cc||cc}
    \hline
    \textbf{Method} & FVD $\downarrow$ & D.D $\uparrow$ & CSD-Score$\uparrow$ & CLIP-T $\uparrow$ & Face Sim $\uparrow$ & DINO $\uparrow$\\
    \hline\hline
    \textbf{Ours} & \textbf{589.8} & \textbf{0.2283} & \textbf{0.427} & \textbf{0.2943} & \textbf{0.548} & \textbf{0.497} \\
    w/o res. & 665.4 & 0.2117 & 0.394 & 0.2855 & 0.497 & 0.457\\
    lightweight   & 655.1 & 0.2090 & 0.408 & 0.2894 & 0.527 & 0.483 \\
    prompt input  & 653.7 & 0.2058 & 0.411 & 0.2942 & 0.529 & 0.495 \\
    \hline
  \end{tabular}
  }
  \caption{Quantitative ablations on different tasks and metrics, including FVD/Dynamic Degree (`D.D.') for Personalized Human Action Video Generation, CSD-Score/CLIP-T for Text-to-Video Stylization, and Face Sim/DINO for Identity-Personalized Image Generation.}
  \label{tab:ablation}
\end{table}

\paragraph{Does the response map effectively guide the network learning?}
To answer this question, we examine two sub-questions:
(1) Does Stage-I accurately predict response maps? $\rightarrow$ On the validation set, Stage-I predictions closely match the ground-truth response maps (cosine sim. 0.77), confirming successful first-stage learning.
(2) Does Stage-II genuinely utilize response maps as guidance? $\rightarrow$ We observe that the response maps of LoRAs predicted by Stage-II remain similar to both Stage-I outputs (cosine sim. 0.91) and ground truth (cosine sim. 0.83), indicating that response information effectively guides Stage-II to identify and emphasize high-response regions.



\paragraph{Alternative way to capture low-level features.}
In Stage-I, we supervise the network using LoRA response maps, encouraging it to learn low-level yet informative patterns that facilitate LoRA prediction. Alternatively, one could employ a \textit{lightweight} hypernetwork at this stage—still predicting LoRA weights—to implicitly bias the model toward learning low-level features, a strategy used in many representation learning principles such as \cite{he2022masked}. As shown in \cref{tab:ablation} “lightweight”, while this alternative improves performance, it still significantly underperforms our proposed approach.

\paragraph{Using base model parameters as input.}
Another key design in our architecture is the use of base model weights as input, combined with user prompts integrated via cross-attention. This approach avoids brute-force prompt-to-LoRA mapping and enables more effective adaptation learning.
As listed in \cref{tab:ablation} “prompt input”, using user prompts as input degrades our performance.

\begin{figure}[t]
    \centering
    \includegraphics[width=0.7\linewidth]{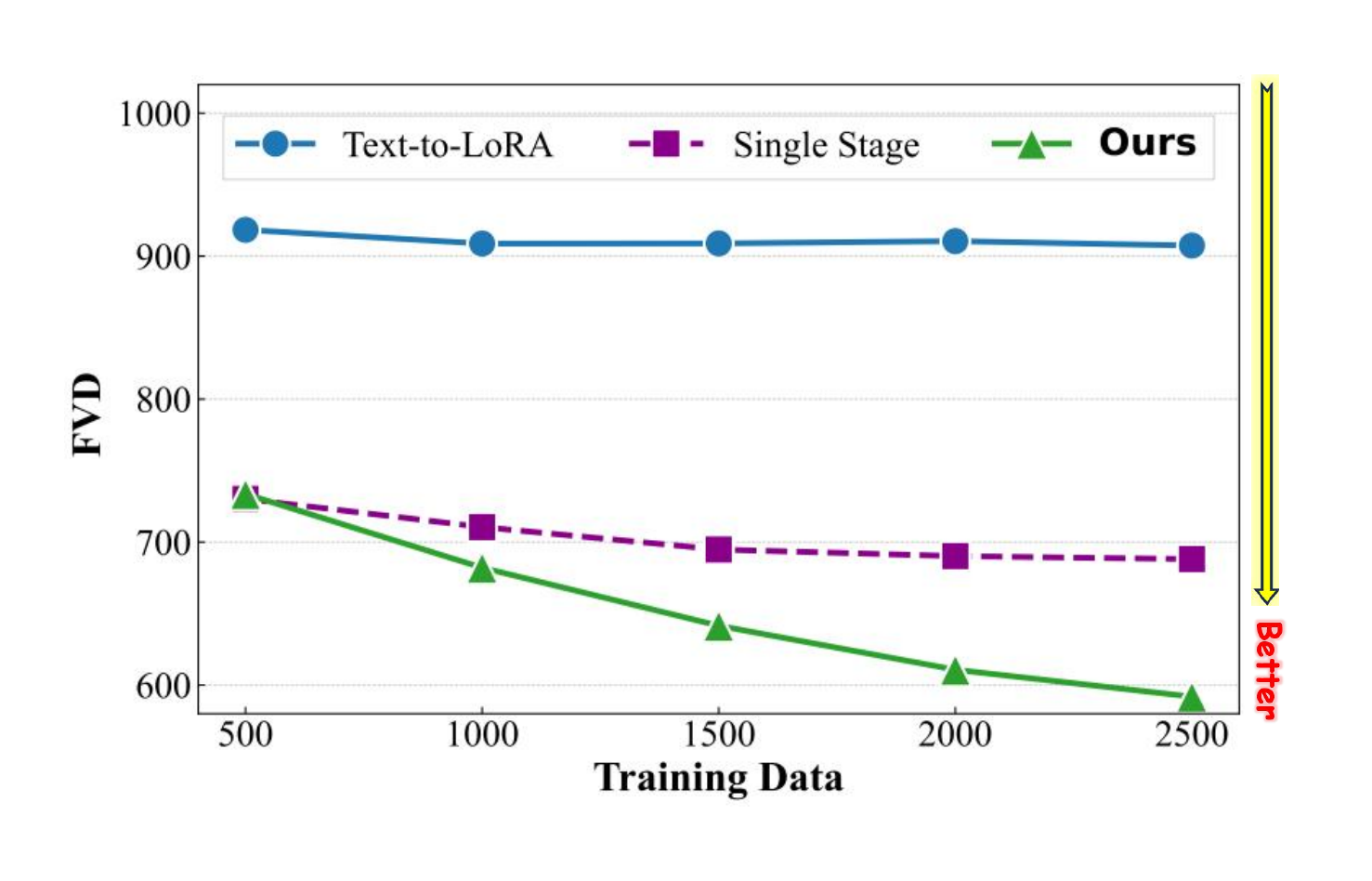}
    \caption{The comparison of scaling up the training data on text-conditioned Human Video Action Generation. Our LoFA consistently benefits from more training data.}
    \label{fig:scale_up}
    \vspace{-0.3cm}
\end{figure}

\paragraph{Scalability.}
Increasing the scale of training data (LoRA pairs) is generally expected to enhance model capacity, yet it also raises learning complexity, especially for the parameter prediction task. 
As shown in \cref{fig:scale_up}, our method consistently benefits from more training data, demonstrating that its design alleviates learning difficulties and maintains stability—enabling effective scaling with more data. In contrast, neither Text-to-LoRA \cite{charakorn2025texttolora} nor “Single Stage” (\ie, without the guidance from response maps) baseline shows comparable gains, likely due to their limited capacity to handle increased learning complexity.

\section{Conclusion}
We presented LoFA, a novel framework for fast and effective personalization of visual generative models. By identifying and leveraging the inherent structures in LoRA response distributions, we designed a two-stage hypernetwork that efficiently predicts full, uncompressed LoRA weights from diverse user inputs. Our approach overcomes key limitations of existing personalization methods—eliminating the need for lengthy optimization while maintaining high output quality. Extensive experiments demonstrate that LoFA achieves comparable or superior results to per-case optimized LoRA, while reducing adaptation time from hours to seconds. 
Looking forward, LoFA establishes a new paradigm for efficient model adaptation that could enable various real-time personalization applications.


{
    \small

    \bibliographystyle{ieeenat_fullname}

\begin{thebibliography}{59}
\providecommand{\natexlab}[1]{#1}
\providecommand{\url}[1]{\texttt{#1}}
\expandafter\ifx\csname urlstyle\endcsname\relax
  \providecommand{\doi}[1]{doi: #1}\else
  \providecommand{\doi}{doi: \begingroup \urlstyle{rm}\Url}\fi

\bibitem[Bai et~al.(2025)Bai, Chen, Liu, Wang, Ge, Song, Dang, Wang, Wang, Tang, et~al.]{bai2025qwen2}
Shuai Bai, Keqin Chen, Xuejing Liu, Jialin Wang, Wenbin Ge, Sibo Song, Kai Dang, Peng Wang, Shijie Wang, Jun Tang, et~al.
\newblock Qwen2. 5-vl technical report.
\newblock \emph{arXiv preprint arXiv:2502.13923}, 2025.

\bibitem[Blattmann et~al.(2023)Blattmann, Dockhorn, Kulal, Mendelevitch, Kilian, Lorenz, Levi, English, Voleti, Letts, et~al.]{stablevideodiffusion}
Andreas Blattmann, Tim Dockhorn, Sumith Kulal, Daniel Mendelevitch, Maciej Kilian, Dominik Lorenz, Yam Levi, Zion English, Vikram Voleti, Adam Letts, et~al.
\newblock Stable video diffusion: Scaling latent video diffusion models to large datasets.
\newblock \emph{arXiv preprint arXiv:2311.15127}, 2023.

\bibitem[Caron et~al.(2021)Caron, Touvron, Misra, J{\'e}gou, Mairal, Bojanowski, and Joulin]{caron2021emerging}
Mathilde Caron, Hugo Touvron, Ishan Misra, Herv{\'e} J{\'e}gou, Julien Mairal, Piotr Bojanowski, and Armand Joulin.
\newblock Emerging properties in self-supervised vision transformers.
\newblock In \emph{ICCV}, 2021.

\bibitem[Charakorn et~al.(2025)Charakorn, Cetin, Tang, and Lange]{charakorn2025texttolora}
Rujikorn Charakorn, Edoardo Cetin, Yujin Tang, and Robert~Tjarko Lange.
\newblock Text-to-lo{RA}: Instant transformer adaption.
\newblock In \emph{ICML}, 2025.

\bibitem[Chung et~al.(2024)Chung, Hou, Longpre, Zoph, Tay, Fedus, Li, Wang, Dehghani, Brahma, et~al.]{chung2024scaling}
Hyung~Won Chung, Le Hou, Shayne Longpre, Barret Zoph, Yi Tay, William Fedus, Yunxuan Li, Xuezhi Wang, Mostafa Dehghani, Siddhartha Brahma, et~al.
\newblock Scaling instruction-finetuned language models.
\newblock \emph{JMLR}, 2024.

\bibitem[Dong et~al.(2022)Dong, Wei, and Lin]{dong2022dreamartist}
Ziyi Dong, Pengxu Wei, and Liang Lin.
\newblock Dreamartist: Towards controllable one-shot text-to-image generation via positive-negative prompt-tuning.
\newblock \emph{arXiv preprint arXiv:2211.11337}, 2022.

\bibitem[Gal et~al.(2022)Gal, Alaluf, Atzmon, Patashnik, Bermano, Chechik, and Cohen-Or]{textualinversion}
Rinon Gal, Yuval Alaluf, Yuval Atzmon, Or Patashnik, Amit~H Bermano, Gal Chechik, and Daniel Cohen-Or.
\newblock An image is worth one word: Personalizing text-to-image generation using textual inversion.
\newblock \emph{arXiv preprint arXiv:2208.01618}, 2022.

\bibitem[Gu et~al.(2025)Gu, Yan, Lu, Li, Dou, Si, Dong, Liu, Lin, Liu, Wang, and Liu]{das}
Zekai Gu, Rui Yan, Jiahao Lu, Peng Li, Zhiyang Dou, Chenyang Si, Zhen Dong, Qifeng Liu, Cheng Lin, Ziwei Liu, Wenping Wang, and Yuan Liu.
\newblock Diffusion as shader: 3d-aware video diffusion for versatile video generation control.
\newblock In \emph{SIGGRAPH}, 2025.

\bibitem[Gururangan et~al.(2020)Gururangan, Marasović, Swayamdipta, Lo, Beltagy, Downey, and Smith]{dontstoppretraining2020}
Suchin Gururangan, Ana Marasović, Swabha Swayamdipta, Kyle Lo, Iz Beltagy, Doug Downey, and Noah~A. Smith.
\newblock Don't stop pretraining: Adapt language models to domains and tasks.
\newblock In \emph{ACL}, 2020.

\bibitem[Ha et~al.(2016)Ha, Dai, and Le]{ha2016hypernetworks}
David Ha, Andrew Dai, and Quoc~V Le.
\newblock Hypernetworks.
\newblock \emph{arXiv preprint arXiv:1609.09106}, 2016.

\bibitem[He et~al.(2022)He, Chen, Xie, Li, Doll{\'a}r, and Girshick]{he2022masked}
Kaiming He, Xinlei Chen, Saining Xie, Yanghao Li, Piotr Doll{\'a}r, and Ross Girshick.
\newblock Masked autoencoders are scalable vision learners.
\newblock In \emph{CVPR}, 2022.

\bibitem[Hessel et~al.(2021)Hessel, Holtzman, Forbes, Le~Bras, and Choi]{hessel2021clipscore}
Jack Hessel, Ari Holtzman, Maxwell Forbes, Ronan Le~Bras, and Yejin Choi.
\newblock Clipscore: A reference-free evaluation metric for image captioning.
\newblock In \emph{EMNLP}, 2021.

\bibitem[Ho et~al.(2020)Ho, Jain, and Abbeel]{ho2020denoising}
Jonathan Ho, Ajay Jain, and Pieter Abbeel.
\newblock Denoising diffusion probabilistic models.
\newblock \emph{NeurIPS}, 2020.

\bibitem[Houlsby et~al.(2019)Houlsby, Giurgiu, Jastrzebski, Morrone, De~Laroussilhe, Gesmundo, Attariyan, and Gelly]{houlsby2019parameter}
Neil Houlsby, Andrei Giurgiu, Stanislaw Jastrzebski, Bruna Morrone, Quentin De~Laroussilhe, Andrea Gesmundo, Mona Attariyan, and Sylvain Gelly.
\newblock Parameter-efficient transfer learning for nlp.
\newblock In \emph{ICML}, 2019.

\bibitem[Hu et~al.(2022)Hu, Shen, Wallis, Allen-Zhu, Li, Wang, Wang, Chen, et~al.]{hu2022lora}
Edward~J Hu, Yelong Shen, Phillip Wallis, Zeyuan Allen-Zhu, Yuanzhi Li, Shean Wang, Lu Wang, Weizhu Chen, et~al.
\newblock Lora: Low-rank adaptation of large language models.
\newblock In \emph{ICLR}, 2022.

\bibitem[Huang et~al.(2024)Huang, He, Yu, Zhang, Si, Jiang, Zhang, Wu, Jin, Chanpaisit, Wang, Chen, Wang, Lin, Qiao, and Liu]{huang2023vbench}
Ziqi Huang, Yinan He, Jiashuo Yu, Fan Zhang, Chenyang Si, Yuming Jiang, Yuanhan Zhang, Tianxing Wu, Qingyang Jin, Nattapol Chanpaisit, Yaohui Wang, Xinyuan Chen, Limin Wang, Dahua Lin, Yu Qiao, and Ziwei Liu.
\newblock {VBench}: Comprehensive benchmark suite for video generative models.
\newblock In \emph{CVPR}, 2024.

\bibitem[Jin et~al.(2024)Jin, Wang, Tang, Zhao, Zhou, Tang, and You]{jin2024conditional}
Xiaolong Jin, Kai Wang, Dongwen Tang, Wangbo Zhao, Yukun Zhou, Junshu Tang, and Yang You.
\newblock Conditional lora parameter generation.
\newblock \emph{arXiv preprint arXiv:2408.01415}, 2024.

\bibitem[Karimi~Mahabadi et~al.(2021)Karimi~Mahabadi, Henderson, and Ruder]{karimi2021compacter}
Rabeeh Karimi~Mahabadi, James Henderson, and Sebastian Ruder.
\newblock Compacter: Efficient low-rank hypercomplex adapter layers.
\newblock In \emph{NeurIPS}, 2021.

\bibitem[Karras et~al.(2019)Karras, Laine, and Aila]{karras2019style}
Tero Karras, Samuli Laine, and Timo Aila.
\newblock A style-based generator architecture for generative adversarial networks.
\newblock In \emph{CVPR}, 2019.

\bibitem[Karras et~al.(2020)Karras, Laine, Aittala, Hellsten, Lehtinen, and Aila]{karras2020analyzing}
Tero Karras, Samuli Laine, Miika Aittala, Janne Hellsten, Jaakko Lehtinen, and Timo Aila.
\newblock Analyzing and improving the image quality of stylegan.
\newblock In \emph{CVPR}, 2020.

\bibitem[Kopiczko et~al.(2024)Kopiczko, Blankevoort, and Asano]{kopiczko2024vera}
Dawid~J Kopiczko, Tijmen Blankevoort, and Yuki~M Asano.
\newblock Vera: Vector-based random matrix adaptation.
\newblock In \emph{ICLR}, 2024.

\bibitem[Kumari et~al.(2023)Kumari, Zhang, Zhang, Shechtman, and Zhu]{customdiffusion}
Nupur Kumari, Bingliang Zhang, Richard Zhang, Eli Shechtman, and Jun-Yan Zhu.
\newblock Multi-concept customization of text-to-image diffusion.
\newblock In \emph{CVPR}, 2023.

\bibitem[Li et~al.(2024)Li, Cao, Wang, Qi, Cheng, and Shan]{li2024photomaker}
Zhen Li, Mingdeng Cao, Xintao Wang, Zhongang Qi, Ming-Ming Cheng, and Ying Shan.
\newblock Photomaker: Customizing realistic human photos via stacked id embedding.
\newblock In \emph{CVPR}, 2024.

\bibitem[Lin et~al.(2023)Lin, Zeng, Lu, Cai, Zhang, Wang, and Zhang]{lin2023motion}
Jing Lin, Ailing Zeng, Shunlin Lu, Yuanhao Cai, Ruimao Zhang, Haoqian Wang, and Lei Zhang.
\newblock Motion-x: A large-scale 3d expressive whole-body human motion dataset.
\newblock \emph{NeurIPS}, 2023.

\bibitem[Lin et~al.(2020)Lin, Madotto, and Fung]{lin2020exploring}
Zhaojiang Lin, Andrea Madotto, and Pascale Fung.
\newblock Exploring versatile generative language model via parameter-efficient transfer learning.
\newblock \emph{arXiv preprint arXiv:2004.03829}, 2020.

\bibitem[Lipman et~al.(2022)Lipman, Chen, Ben-Hamu, Nickel, and Le]{lipman2022flow}
Yaron Lipman, Ricky~TQ Chen, Heli Ben-Hamu, Maximilian Nickel, and Matt Le.
\newblock Flow matching for generative modeling.
\newblock \emph{arXiv preprint arXiv:2210.02747}, 2022.

\bibitem[Loshchilov and Hutter(2017)]{loshchilov2017decoupled}
Ilya Loshchilov and Frank Hutter.
\newblock Decoupled weight decay regularization.
\newblock \emph{arXiv preprint arXiv:1711.05101}, 2017.

\bibitem[Nan et~al.(2024)Nan, Xie, Zhou, Fan, Yang, Chen, Li, Yang, and Tai]{nan2024openvid}
Kepan Nan, Rui Xie, Penghao Zhou, Tiehan Fan, Zhenheng Yang, Zhijie Chen, Xiang Li, Jian Yang, and Ying Tai.
\newblock Openvid-1m: A large-scale high-quality dataset for text-to-video generation.
\newblock \emph{arXiv preprint arXiv:2407.02371}, 2024.

\bibitem[Patashnik et~al.(2025)Patashnik, Parmar, Rao, Kara, Caba~Heilbron, Cohen-Or, Matthew~Rehg, and Zhu]{visaulart}
Or Patashnik, Gaurav Parmar, Anyi Rao, Ozgur Kara, Fabian Caba~Heilbron, Daniel Cohen-Or, James Matthew~Rehg, and Jun-Yan Zhu.
\newblock Ai for creative visual content generation, editing and understanding.
\newblock In \emph{SIGGRAPH}, 2025.

\bibitem[Peebles et~al.(2022)Peebles, Radosavovic, Brooks, Efros, and Malik]{peebles2022learning}
William Peebles, Ilija Radosavovic, Tim Brooks, Alexei~A Efros, and Jitendra Malik.
\newblock Learning to learn with generative models of neural network checkpoints.
\newblock \emph{arXiv preprint arXiv:2209.12892}, 2022.

\bibitem[Podell et~al.(2024)Podell, English, Lacey, Blattmann, Dockhorn, M{\"u}ller, Penna, and Rombach]{podell2024sdxl}
Dustin Podell, Zion English, Kyle Lacey, Andreas Blattmann, Tim Dockhorn, Jonas M{\"u}ller, Joe Penna, and Robin Rombach.
\newblock {SDXL}: Improving latent diffusion models for high-resolution image synthesis.
\newblock In \emph{ICLR}, 2024.

\bibitem[Radford et~al.(2021)Radford, Kim, Hallacy, Ramesh, Goh, Agarwal, Sastry, Askell, Mishkin, Clark, et~al.]{radford2021learning}
Alec Radford, Jong~Wook Kim, Chris Hallacy, Aditya Ramesh, Gabriel Goh, Sandhini Agarwal, Girish Sastry, Amanda Askell, Pamela Mishkin, Jack Clark, et~al.
\newblock Learning transferable visual models from natural language supervision.
\newblock In \emph{ICML}, 2021.

\bibitem[Rebuffi et~al.(2017)Rebuffi, Bilen, and Vedaldi]{rebuffi2017learning}
Sylvestre-Alvise Rebuffi, Hakan Bilen, and Andrea Vedaldi.
\newblock Learning multiple visual domains with residual adapters.
\newblock In \emph{NeurIPS}, 2017.

\bibitem[Ren et~al.(2025)Ren, Shen, Huang, Ling, Lu, Nimier-David, M\"uller, Keller, Fidler, and Gao]{ren2025gen3c}
Xuanchi Ren, Tianchang Shen, Jiahui Huang, Huan Ling, Yifan Lu, Merlin Nimier-David, Thomas M\"uller, Alexander Keller, Sanja Fidler, and Jun Gao.
\newblock Gen3c: 3d-informed world-consistent video generation with precise camera control.
\newblock In \emph{CVPR}, 2025.

\bibitem[Rombach et~al.(2022)Rombach, Blattmann, Lorenz, Esser, and Ommer]{stablediffusion}
Robin Rombach, Andreas Blattmann, Dominik Lorenz, Patrick Esser, and Bjorn Ommer.
\newblock { High-Resolution Image Synthesis with Latent Diffusion Models }.
\newblock In \emph{CVPR}, 2022.

\bibitem[Ruiz et~al.(2023)Ruiz, Li, Jampani, Pritch, Rubinstein, and Aberman]{ruiz2023dreambooth}
Nataniel Ruiz, Yuanzhen Li, Varun Jampani, Yael Pritch, Michael Rubinstein, and Kfir Aberman.
\newblock Dreambooth: Fine tuning text-to-image diffusion models for subject-driven generation.
\newblock In \emph{CVPR}, 2023.

\bibitem[Ruiz et~al.(2024)Ruiz, Li, Jampani, Wei, Hou, Pritch, Wadhwa, Rubinstein, and Aberman]{hyperdreambooth}
Nataniel Ruiz, Yuanzhen Li, Varun Jampani, Wei Wei, Tingbo Hou, Yael Pritch, Neal Wadhwa, Michael Rubinstein, and Kfir Aberman.
\newblock { HyperDreamBooth: HyperNetworks for Fast Personalization of Text-to-Image Models =}.
\newblock In \emph{CVPR}, 2024.

\bibitem[Saharia et~al.(2022)Saharia, Chan, Saxena, Li, Whang, Denton, Ghasemipour, Gontijo~Lopes, Karagol~Ayan, Salimans, et~al.]{saharia2022photorealistic}
Chitwan Saharia, William Chan, Saurabh Saxena, Lala Li, Jay Whang, Emily~L Denton, Kamyar Ghasemipour, Raphael Gontijo~Lopes, Burcu Karagol~Ayan, Tim Salimans, et~al.
\newblock Photorealistic text-to-image diffusion models with deep language understanding.
\newblock In \emph{NeurIPS}, 2022.

\bibitem[Shao et~al.(2025)Shao, Yan, Liu, Chen, Chen, Long, Yan, Li, Zhang, Sebe, et~al.]{incontext}
Yihua Shao, Minxi Yan, Yang Liu, Siyu Chen, Wenjie Chen, Xinwei Long, Ziyang Yan, Lei Li, Chenyu Zhang, Nicu Sebe, et~al.
\newblock In-context meta lora generation.
\newblock \emph{arXiv preprint arXiv:2501.17635}, 2025.

\bibitem[Shenaj et~al.(2025)Shenaj, Bohdal, Ozay, Zanuttigh, and Michieli]{lorarar}
Donald Shenaj, Ondrej Bohdal, Mete Ozay, Pietro Zanuttigh, and Umberto Michieli.
\newblock Lora. rar: Learning to merge loras via hypernetworks for subject-style conditioned image generation.
\newblock In \emph{ICCV}, 2025.

\bibitem[Shi et~al.(2024)Shi, Huang, Wang, Bian, Li, Zhang, Zhang, Cheung, See, Qin, et~al.]{shi2024motion}
Xiaoyu Shi, Zhaoyang Huang, Fu-Yun Wang, Weikang Bian, Dasong Li, Yi Zhang, Manyuan Zhang, Ka~Chun Cheung, Simon See, Hongwei Qin, et~al.
\newblock Motion-i2v: Consistent and controllable image-to-video generation with explicit motion modeling.
\newblock \emph{SIGGRAPH}, 2024.

\bibitem[Singer et~al.(2022)Singer, Polyak, Hayes, Yin, An, Zhang, Hu, Yang, Ashual, Gafni, et~al.]{singer2022make}
Uriel Singer, Adam Polyak, Thomas Hayes, Xi Yin, Jie An, Songyang Zhang, Qiyuan Hu, Harry Yang, Oron Ashual, Oran Gafni, et~al.
\newblock Make-a-video: Text-to-video generation without text-video data.
\newblock \emph{arXiv preprint arXiv:2209.14792}, 2022.

\bibitem[Somepalli et~al.(2024)Somepalli, Gupta, Gupta, Palta, Goldblum, Geiping, Shrivastava, and Goldstein]{somepalli2024measuring}
Gowthami Somepalli, Anubhav Gupta, Kamal Gupta, Shramay Palta, Micah Goldblum, Jonas Geiping, Abhinav Shrivastava, and Tom Goldstein.
\newblock Measuring style similarity in diffusion models.
\newblock \emph{arXiv preprint arXiv:2404.01292}, 2024.

\bibitem[Unterthiner et~al.(2018)Unterthiner, Van~Steenkiste, Kurach, Marinier, Michalski, and Gelly]{unterthiner2018towards}
Thomas Unterthiner, Sjoerd Van~Steenkiste, Karol Kurach, Raphael Marinier, Marcin Michalski, and Sylvain Gelly.
\newblock Towards accurate generative models of video: A new metric \& challenges.
\newblock \emph{arXiv preprint arXiv:1812.01717}, 2018.

\bibitem[Vaswani et~al.(2017)Vaswani, Shazeer, Parmar, Uszkoreit, Jones, Gomez, Kaiser, and Polosukhin]{attention}
Ashish Vaswani, Noam Shazeer, Niki Parmar, Jakob Uszkoreit, Llion Jones, Aidan~N Gomez, \L~ukasz Kaiser, and Illia Polosukhin.
\newblock Attention is all you need.
\newblock In \emph{NeurIPS}, 2017.

\bibitem[Wan et~al.(2025)Wan, Wang, Ai, Wen, Mao, Xie, Chen, Yu, Zhao, Yang, et~al.]{wan2025wan}
Team Wan, Ang Wang, Baole Ai, Bin Wen, Chaojie Mao, Chen-Wei Xie, Di Chen, Feiwu Yu, Haiming Zhao, Jianxiao Yang, et~al.
\newblock Wan: Open and advanced large-scale video generative models.
\newblock \emph{arXiv preprint arXiv:2503.20314}, 2025.

\bibitem[Wang et~al.(2024)Wang, Tang, Zeng, Yin, Xu, Zhou, Zang, Darrell, Liu, and You]{wang2024neural}
Kai Wang, Dongwen Tang, Boya Zeng, Yida Yin, Zhaopan Xu, Yukun Zhou, Zelin Zang, Trevor Darrell, Zhuang Liu, and Yang You.
\newblock Neural network diffusion.
\newblock \emph{arXiv preprint arXiv:2402.13144}, 2024.

\bibitem[Wang et~al.(2025{\natexlab{a}})Wang, Tang, Zhao, Sch{\"u}rholt, Wang, and You]{wangscaling}
Kai Wang, Dongwen Tang, Wangbo Zhao, Konstantin Sch{\"u}rholt, Zhangyang Wang, and Yang You.
\newblock Scaling up parameter generation: A recurrent diffusion approach.
\newblock In \emph{NeurIPS}, 2025{\natexlab{a}}.

\bibitem[Wang et~al.(2025{\natexlab{b}})Wang, Zhang, Gao, Wang, Zhou, Zhang, Yan, and Sang]{wang2025unianimate}
Xiang Wang, Shiwei Zhang, Changxin Gao, Jiayu Wang, Xiaoqiang Zhou, Yingya Zhang, Luxin Yan, and Nong Sang.
\newblock Unianimate: Taming unified video diffusion models for consistent human image animation.
\newblock \emph{Science China Information Sciences}, 2025{\natexlab{b}}.

\bibitem[Wei et~al.(2021)Wei, Bosma, Zhao, Guu, Yu, Lester, Du, Dai, and Le]{wei2021finetuned}
Jason Wei, Maarten Bosma, Vincent~Y Zhao, Kelvin Guu, Adams~Wei Yu, Brian Lester, Nan Du, Andrew~M Dai, and Quoc~V Le.
\newblock Finetuned language models are zero-shot learners.
\newblock \emph{arXiv preprint arXiv:2109.01652}, 2021.

\bibitem[Wu et~al.(2024)Wu, Shi, Wei, Sun, Yang, and Shen]{wu2024difflora}
Yujia Wu, Yiming Shi, Jiwei Wei, Chengwei Sun, Yang Yang, and Heng~Tao Shen.
\newblock Difflora: Generating personalized low-rank adaptation weights with diffusion.
\newblock \emph{arXiv preprint arXiv:2408.06740}, 2024.

\bibitem[Yang et~al.(2023)Yang, Zeng, Yuan, and Li]{yang2023effective}
Zhendong Yang, Ailing Zeng, Chun Yuan, and Yu Li.
\newblock Effective whole-body pose estimation with two-stages distillation.
\newblock In \emph{ICCV}, 2023.

\bibitem[Ye et~al.(2023)Ye, Zhang, Liu, Han, and Yang]{ye2023ip}
Hu Ye, Jun Zhang, Sibo Liu, Xiao Han, and Wei Yang.
\newblock Ip-adapter: Text compatible image prompt adapter for text-to-image diffusion models.
\newblock \emph{arXiv preprint arXiv:2308.06721}, 2023.

\bibitem[Ye et~al.(2025)Ye, Huang, Wang, Wan, Zhang, and Luo]{ye2025stylemaster}
Zixuan Ye, Huijuan Huang, Xintao Wang, Pengfei Wan, Di Zhang, and Wenhan Luo.
\newblock Stylemaster: Stylize your video with artistic generation and translation.
\newblock In \emph{CVPR}, 2025.

\bibitem[YU et~al.(2025)YU, Hu, Xing, and Shan]{mark2025trajectorycrafter}
Mark YU, Wenbo Hu, Jinbo Xing, and Ying Shan.
\newblock Trajectorycrafter: Redirecting camera trajectory for monocular videos via diffusion models.
\newblock In \emph{ICCV}, 2025.

\bibitem[Zhang et~al.(2025{\natexlab{a}})Zhang, Paiss, Zada, Karnad, Jacobs, Pritch, Mosseri, Shou, Wadhwa, and Ruiz]{zhang2025recapture}
David~Junhao Zhang, Roni Paiss, Shiran Zada, Nikhil Karnad, David~E Jacobs, Yael Pritch, Inbar Mosseri, Mike~Zheng Shou, Neal Wadhwa, and Nataniel Ruiz.
\newblock Recapture: Generative video camera controls for user-provided videos using masked video fine-tuning.
\newblock In \emph{CVPR}, 2025{\natexlab{a}}.

\bibitem[Zhang et~al.(2023{\natexlab{a}})Zhang, Rao, and Agrawala]{zhang2023adding}
Lvmin Zhang, Anyi Rao, and Maneesh Agrawala.
\newblock Adding conditional control to text-to-image diffusion models.
\newblock In \emph{ICCV}, 2023{\natexlab{a}}.

\bibitem[Zhang et~al.(2023{\natexlab{b}})Zhang, Chen, Bukharin, He, Cheng, Chen, and Zhao]{zhang2023adaptive}
Qingru Zhang, Minshuo Chen, Alexander Bukharin, Pengcheng He, Yu Cheng, Weizhu Chen, and Tuo Zhao.
\newblock Adaptive budget allocation for parameter-efficient fine-tuning.
\newblock In \emph{ICLR}, 2023{\natexlab{b}}.

\bibitem[Zhang et~al.(2025{\natexlab{b}})Zhang, Lin, Zeng, Wu, Lu, Fu, Cai, Zhang, Wang, and Zhang]{zhang2025motion}
Yuhong Zhang, Jing Lin, Ailing Zeng, Guanlin Wu, Shunlin Lu, Yurong Fu, Yuanhao Cai, Ruimao Zhang, Haoqian Wang, and Lei Zhang.
\newblock Motion-x++: A large-scale multimodal 3d whole-body human motion dataset.
\newblock \emph{arXiv preprint arXiv:2501.05098}, 2025{\natexlab{b}}.

\end{thebibliography}
}

\clearpage
\setcounter{page}{1}
\maketitlesupplementary

\section{More Results}
\begin{figure*}[t]
    \centering
    \includegraphics[width=1.0\linewidth]{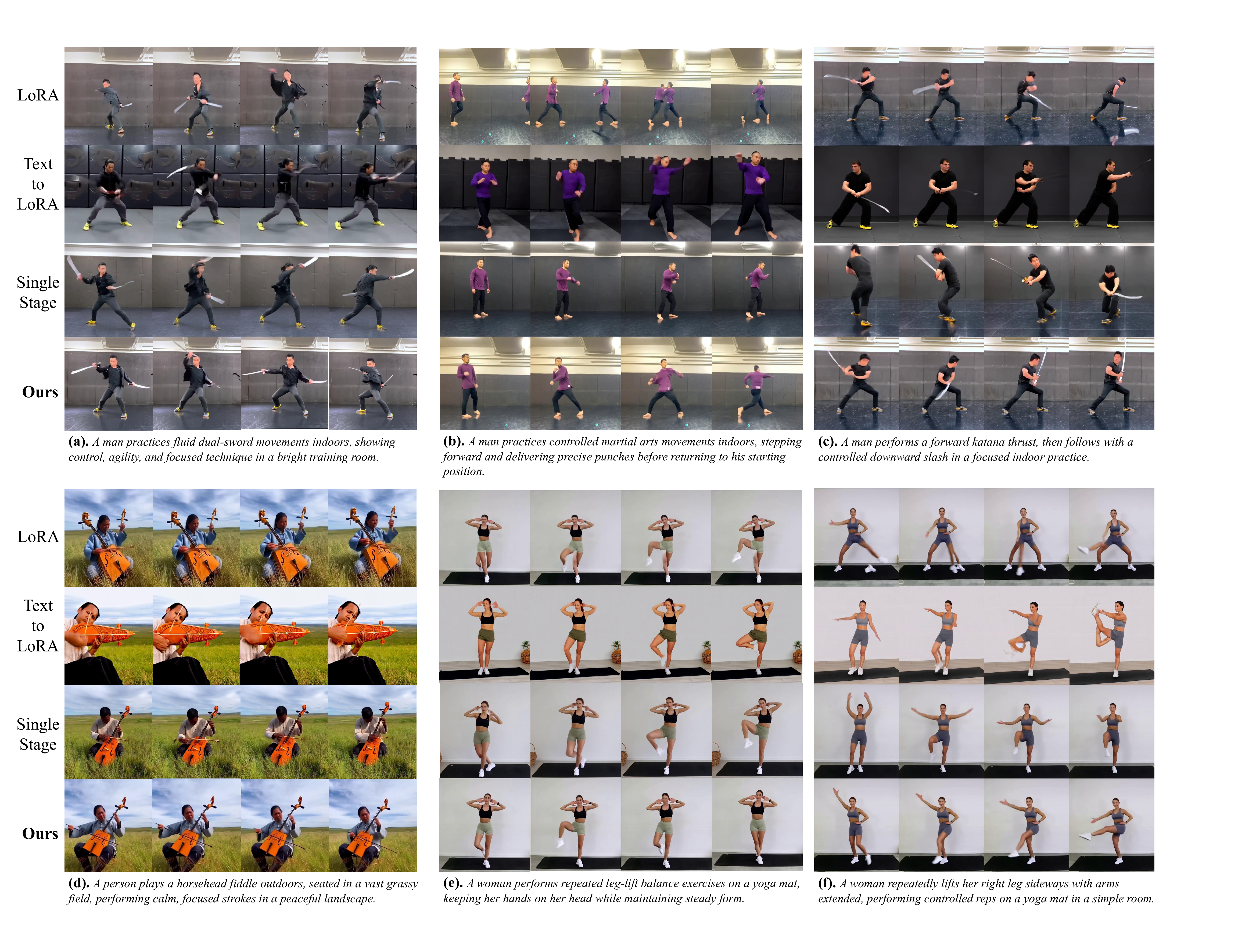}
    \caption{More qualitative results on \textbf{text-conditioned Personalized Human Action Video Generation}.}
    \label{fig:supp_text_cond_result}
\end{figure*}

\begin{figure*}[t]
    \centering
    \includegraphics[width=1.0\linewidth]{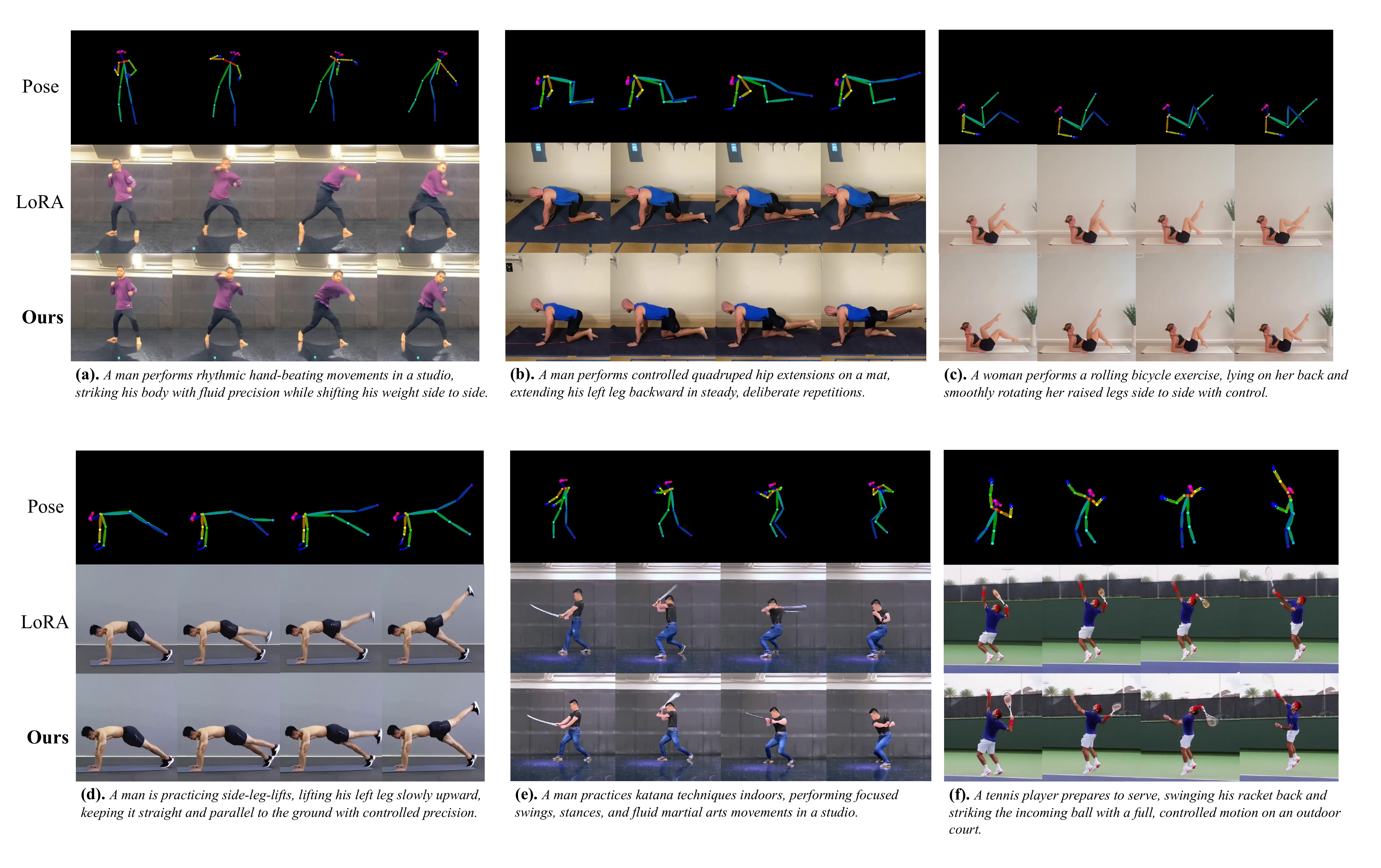}
    \caption{More qualitative results on \textbf{pose-conditioned Personalized Human Action Video Generation}.}
    \label{fig:supp_pose_cond_result}
\end{figure*}

\begin{figure*}[t]
    \centering
    \includegraphics[width=1.0\linewidth]{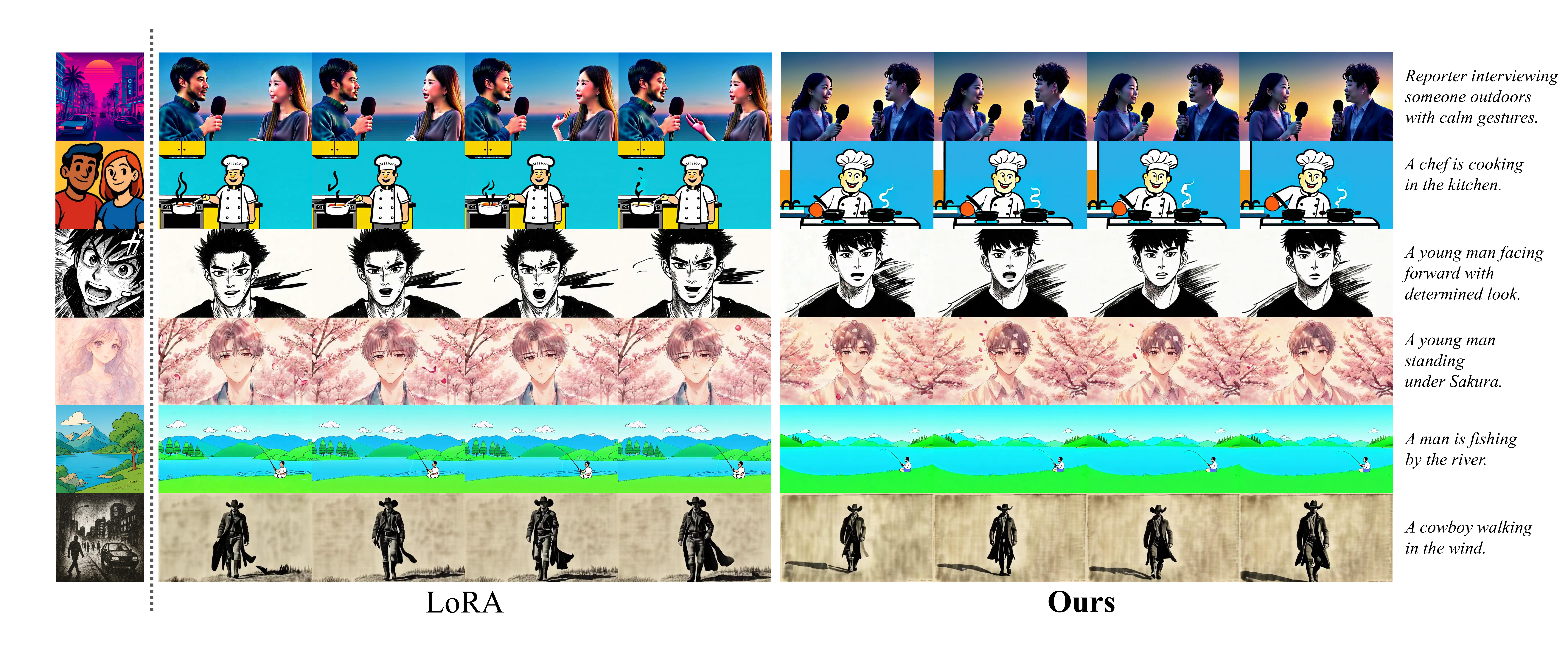}
    \caption{More qualitative results on \textbf{Text-to-Video Stylization}.}
    \label{fig:supp_t2v_stylization}
\end{figure*}

\begin{figure*}[t]
  \centering
  \includegraphics[width=0.9\linewidth]{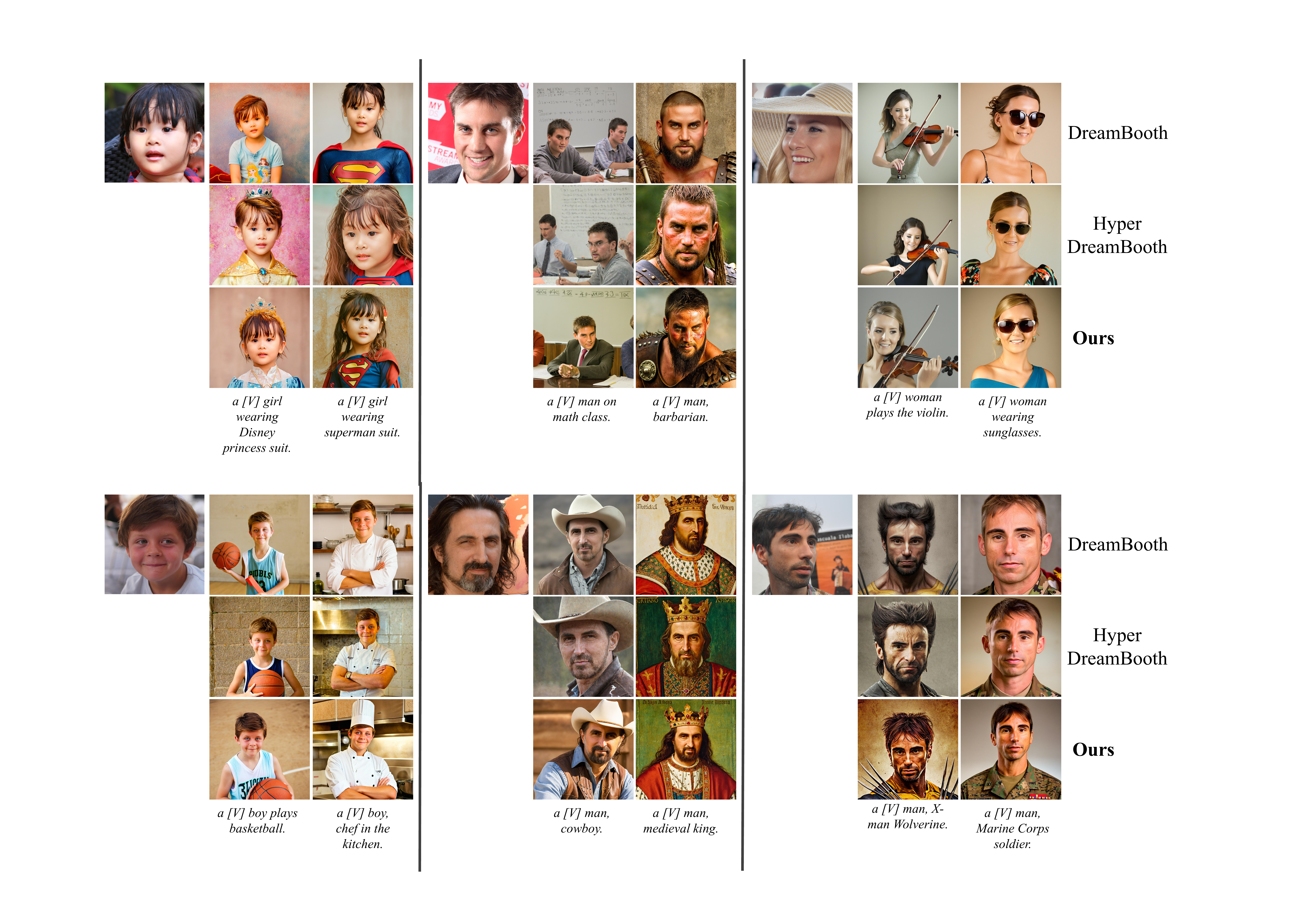}
  \caption{More qualitative results on \textbf{Identity-Personalized Image Generation}.}
  \label{fig:supp_face_compare}
\end{figure*}

\cref{fig:supp_text_cond_result}, \cref{fig:supp_pose_cond_result}, \cref{fig:supp_t2v_stylization} and \cref{fig:supp_face_compare} present more qualitative results on text/pose-conditioned Personalized Human Action Video Generation, Text-to-Video Stylization and Identity-Personalized Image Generation tasks, where our method consistently outperforms other baselines.

We further provide a \textit{\textbf{demo}} video in the supplementary file, named {\textit{demo.mp4}}.
In this video, LoRA \cite{hu2022lora} and its problems are introduced from start to 00:43. Next, the advantage of our LoFA is briefly illustrated. Finally, the qualitative comparisons across all downstream tasks, including the generated videos and images, are shown from 01:03 to the end.

\section{User Study}
To complement our evaluation, we conducted a subjective user study across all personalization tasks. We recruited 50 participants with no expertise in visual generation via an online questionnaire. For each task, we randomly select 5 prompts from the validation set and send them to each method to generate corresponding results, which are displayed to participants side-by-side in a random order alongside the original prompt. During the evaluation, we instruct subjects to select the best result based on prompt alignment and visual quality. As shown in \cref{tab:user_study}, our method achieved a higher win rate across all tasks, demonstrating its clear superiority.

\begin{table*}[t]
\centering
\setlength{\tabcolsep}{4pt}         
\renewcommand{\arraystretch}{1.05}

\resizebox{1.0\linewidth}{!}{
\begin{tabular}{c|ccc|cc|cc|ccc}
\toprule
& \multicolumn{3}{c|}{\small\shortstack{\textit{Text-Cond}\\\textit{Human Action Video}}}
& \multicolumn{2}{c|}{\small\shortstack{\textit{Pose-Cond}\\\textit{Human Action Video}}}
& \multicolumn{2}{c|}{\small\shortstack{\textit{Text-to-Video}\\\textit{Stylization}}}
& \multicolumn{3}{c}{\small\shortstack{\textit{Identity-Personalized}\\\textit{Image Generation}}}
\\
\midrule
& LoRA \cite{hu2022lora} & Text-to-LoRA \cite{charakorn2025texttolora} & \textbf{Ours} 
& LoRA \cite{hu2022lora} & \textbf{Ours}
& LoRA \cite{hu2022lora} & \textbf{Ours}
& DreamBooth \cite{ruiz2023dreambooth} & HyperDreamBooth \cite{hyperdreambooth} & \textbf{Ours} \\
\midrule
Win Rate (\%) 
& 40.8 & 2.4 & \textbf{56.8}
& 48.8 & \textbf{51.2}
& 47.6 & \textbf{52.4}
& 15.2 & 36.8 & \textbf{48.0} \\
\bottomrule
\end{tabular}
}
\caption{User studies across all personalization tasks.}
\label{tab:user_study}
\end{table*}

\section{More Ablations and Analysis}

\begin{figure}[t]
  \centering
  \includegraphics[width=0.8\linewidth]{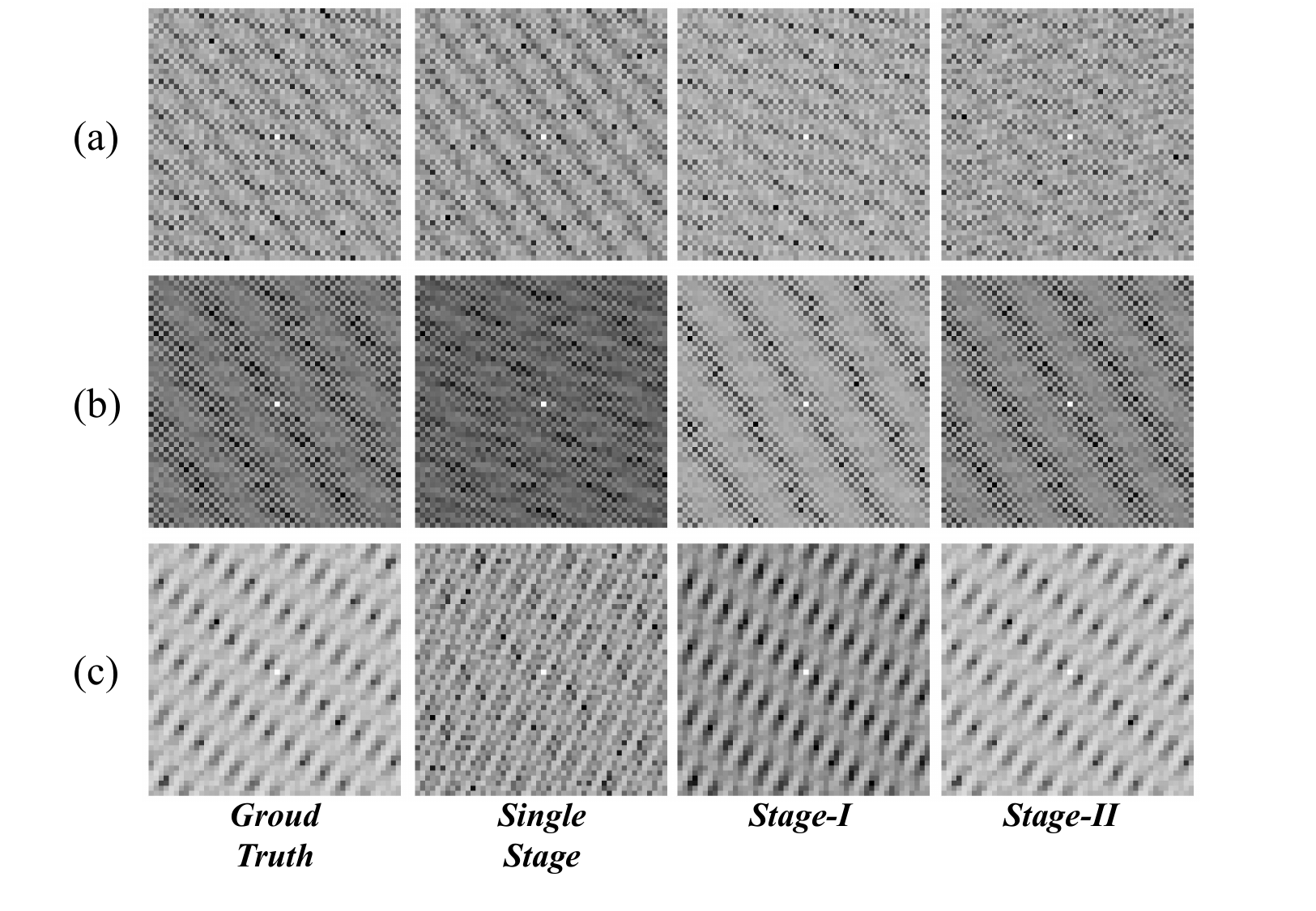}
  \caption{\textbf{Qualitative comparison of LoRA response maps}. Each row corresponds to a distinct task-specific LoRA, while columns represent the response map produced by different models or ground truth.}
  \label{fig:response_map_supp}
\end{figure}

\paragraph{Qualitative comparison of LoRA response maps.}
\cref{fig:response_map_supp} visualizes the response maps derived from the ground truth (GT) LoRA \cite{hu2022lora}, the single-stage prediction (\ie, directly predicting LoRA weights without the guidance from response maps), and both stages of our proposed method. Compared to the GT LoRA, the single-stage prediction's response map shows poor alignment. In contrast, both our Stage-I prediction and the final Stage-II prediction align closely with the GT. This demonstrates that our two-stage learning effectively guides the hypernetwork to identify and emphasize the regions requiring greater attention.

\begin{table}[htbp]
\centering
\small
\setlength{\tabcolsep}{5pt}
\renewcommand{\arraystretch}{1.1}
\resizebox{0.6\linewidth}{!}{
\begin{tabular}{c|c|c|c}
\toprule
Threshold & FVD $\downarrow$ & Clip-T $\uparrow$ & D.D $\uparrow$ \\
\midrule
0.01        & 634.7 & 0.3687 & 0.2219 \\
0.015        & 602.9 & 0.3691 & 0.2245 \\
\textbf{0.02 (ours)} & \textbf{589.8} & \textbf{0.3719} & \textbf{0.2283} \\
0.025        & 597.1 & 0.3712 & 0.2253 \\
0.03        & 622.5 & 0.3620 & 0.2197 \\
\bottomrule
\end{tabular}
}
\caption{Quantitative ablations on the threshold to get the response map, on text-conditioned Personalized
Human Action Video Generation.}
\label{tab:supp_threshold}
\end{table}

\paragraph{Threshold to get the LoRA response map.}

\cref{tab:supp_threshold} ablates the threshold to get the binary-masked LoRA response map. A low threshold (\eg, 0.01) masks a negligible area, revealing no distinct pattern, while a high threshold (\eg, 0.03) masks an excessive area. We find that a threshold between 0.015 and 0.025 provides an optimal balance, yielding stable and high-quality results.

\begin{figure*}[t]
  \centering
  \includegraphics[width=0.85\linewidth]{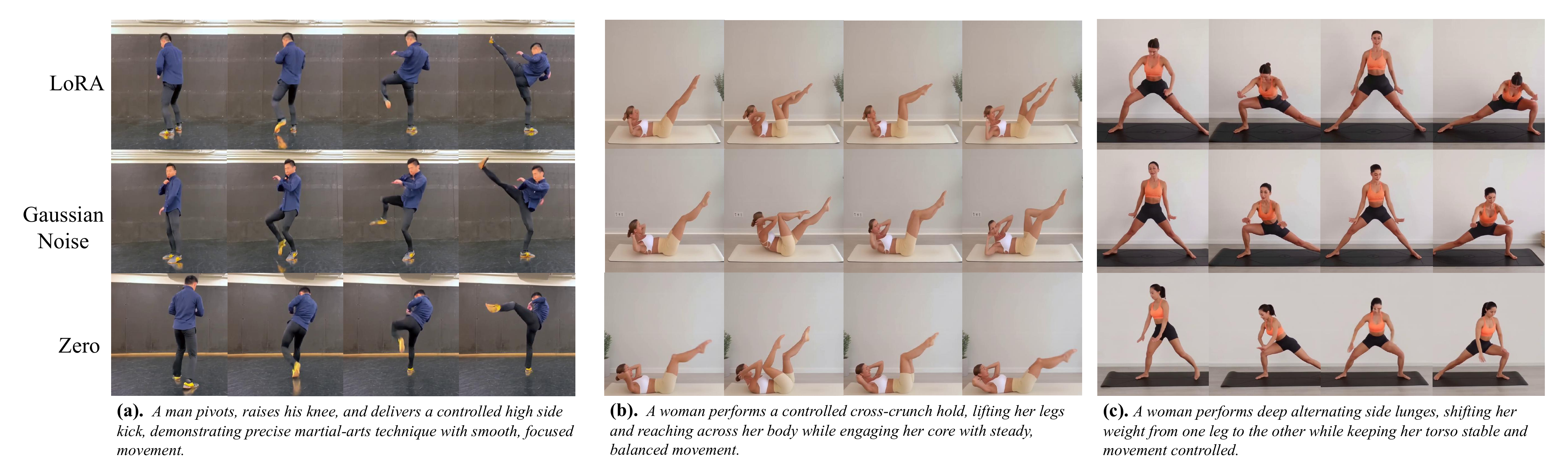}
  \caption{We investigate whether the parameters masked by our threshold are functionally useless. We perturb these parameters by adding Gaussian noise (\textbf{middle}) or setting them to zero (\textbf{bottom}). Compared with the original LoRA (\textbf{Top}), these perturbations cause \textit{no} essential performance difference, confirming that the masked parameters are indeed negligible.}
  \label{fig:supp_noisy_lora}
\end{figure*}

\paragraph{Are the masked parameters truly negligible?}
We investigate whether the parameters masked by our threshold are functionally useless. To test this, we perturb these parameters by adding Gaussian noise or setting them to zero. As shown in \cref{fig:supp_noisy_lora}, these perturbations cause \textit{no} essential performance difference, confirming that the masked parameters are indeed negligible.

\paragraph{Attending to the feature of the Stage-I model.} 
\begin{table}[htbp]
  \centering
  \vspace{2mm}
  \resizebox{0.9\linewidth}{!}{
  \begin{tabular}{c||c|c|c|c}
    \hline
    $N$ Layers & 2 & 4 & 6 & 8 \\
    \hline\hline
    FVD $\downarrow$ & \textbf{660.1} & 668.3 & 678.2 & 680.9 \\
    \hline
    Dynamic Degree $\uparrow$ & \textbf{0.2067} & 0.2051 & 0.1913 & 0.1845 \\
    \hline
  \end{tabular}
  }
  \caption{Quantitative ablations on attending the Stage-I feature on the \textit{first} $N$-th layers in Stage-II, on text-conditioned Personalized Human Action Video Generation.}
  \label{tab:ablation_layer_num}
\end{table}
\begin{table}[htbp]
  \centering
  \small
  \vspace{2mm}
  \resizebox{0.8\linewidth}{!}{
  \begin{tabular}{l||c|c}
    \hline
    Injected Layers & FVD $\downarrow$ & Dynamic Degree $\uparrow$ \\
    \hline\hline
    1st \& 2nd & 660.1 & 0.2067 \\
    2nd \& 4th & 644.3 & 0.2104 \\
    4th \& 6th & 612.2 & 0.2236 \\
    6th \& 8th & 633.9 & 0.2106 \\
    4th \& 8th (ours) & \textbf{589.8} & \textbf{0.2283} \\
    \hline
  \end{tabular}
  }
  \caption{Quantitative ablations on different cross-attention layers for injecting the Stage-I feature, on text-conditioned Personalized Human Action Video Generation.}
  \label{tab:ablation_indexes}
\end{table}

In our Stage-II model, an additional cross-attention layer is introduced to attend to the final layer feature representation of the Stage-I model.
We further conduct an ablation study on \textit{how many layers} and \textit{which layers} should be attended. First, \cref{tab:ablation_layer_num} shows the results of attending the Stage-I feature at the \textit{first} $N$-th layers in Stage-II, indicating that attending at the first $2$-th layers yields the best result. Furthermore, \cref{tab:ablation_indexes} shows that injecting the features into the 4th and 8th blocks yields the best overall performance.



\section{Implementation Details}
\paragraph{Training strategy.}
All experiments follow the same training protocol to train our framework: the first stage is trained for 4,000 steps with a learning rate of 1e‑4, and the second stage for 7,000 steps with a learning rate of 4e‑5. We set loss weights $\lambda_{\text{recon}} = 5$ and $\lambda_{\text{diff}} = 1$, use the default flow-matching objective used in the base model \cite{wan2025wan}, and apply the AdamW \cite{loshchilov2017decoupled} optimizer with 1,000 warm-up steps and a batch size of 4. We use the LoRA parameters from linear projections in attention layers as the training supervision.
During the data preparation, all LoRAs \cite{hu2022lora} are optimized for 1,000 steps with a constant learning rate of 1e-4 and a batch size of 4.

\paragraph{Text prompts.}
\cref{tab:supp_prompts} lists some samples of the text to prompt text-to-video tasks. It shows the \textit{fine-grained} property of the prompts used to predict LoRA weights.

\section{Limitation and Future Work.}
A key limitation of our method is that handling different domain-specific prompts—such as those for human actions, identities, or artistic styles—currently requires training separate networks. The ideal solution is a \textit{single, unified} hypernetwork with strong zero-shot capabilities. Given that our architecture has demonstrated promising scalability, we are confident that scaling up the quantity and diversity of training data will enable the development of such a model in the future.

\begin{table}[t]
\centering
\small
\setlength{\tabcolsep}{8pt}
\renewcommand{\arraystretch}{1.35}

\resizebox{1.0\linewidth}{!}{
\begin{tabular}{|p{0.7\textwidth}|}
\hline
\multicolumn{1}{|c|}{\textbf{Input Text}} \\ \hline

\textbf{Prompt (a):} \\
A woman is performing a workout routine on a black yoga mat placed on a gray floor.  The woman starts by lying on her back with her legs extended straight out. She then lifts one leg at a time, keeping it straight and tapping her toes. She repeats this movement several times, alternating between lifting each leg. The video captures her form and technique as she performs the exercise.
\\ \hline

\textbf{Prompt (b):} \\
In this video, a man is seen practicing with a katana in an indoor studio setting.  His posture is dynamic and focused, indicating that he is engaged in a serious practice session. He begins by standing with his back to the camera, holding the katana in a ready position. As the video progresses, he transitions into various stances and movements, demonstrating different techniques and maneuvers with the sword. His movements are fluid and precise, showcasing his skill and control over the weapon.Throughout the video, the man maintains a strong and confident stance, occasionally shifting his weight from one foot to another while maintaining a firm grip on the katana. His facial expression remains neutral, suggesting concentration and dedication to his practice. 
\\ \hline

\textbf{Prompt (c):} \\
In this video, a man is performing a series of dynamic movements in an indoor studio setting. He begins by standing with his feet apart, arms outstretched to the sides. He then transitions into a series of fluid, athletic movements. His right leg kicks forward while he maintains a balanced stance, showcasing agility and control. After the kick, he smoothly transitions back to a standing position, repeating the sequence with his left leg. Throughout the performance, the man's movements are precise and deliberate, emphasizing the fluidity and grace of the kick walking technique. His body language conveys strength and precision, highlighting the skill involved in executing such a complex movement.
\\ \hline
\end{tabular}
}
\caption{Samples of the detailed input prompts.}
\label{tab:supp_prompts}
\end{table}

%

\end{document}